\newcommand{\negation}{\textbf{negation}}

\newcommand{\reinforcement}{\textbf{reinforcement}}

\newcommand{\lminput}[1]{``\textsl{#1}''}

\documentclass{article} 
\usepackage{arxiv}
\usepackage[T1]{fontenc}
\usepackage{graphicx}
\usepackage{microtype}
\usepackage{hyperref}
\usepackage{url}
\usepackage{booktabs}
\usepackage{caption}
\usepackage{amsmath}
\usepackage{soul}
\usepackage{lineno}
\usepackage{tabularx}
\usepackage{natbib}
\usepackage{bm}
\usepackage{svg}
\usepackage{subcaption}
\definecolor{darkblue}{rgb}{0, 0, 0.5}
\hypersetup{colorlinks=true, citecolor=darkblue, linkcolor=darkblue, urlcolor=darkblue}

\title{From Indirect Object Identification to Syllogisms: Exploring Binary Mechanisms in Transformer Circuits}

\author{
Karim Saraipour \\
University of California, Los Angeles \\
\thanks{Correspondence: karimsaraipour@cs.ucla.edu}
\And
Shichang Zhang \\
Harvard University}

\date{}

\begin{document}


\maketitle

\begin{abstract}
Transformer-based language models (LMs) can perform a wide range of tasks, and mechanistic interpretability (MI) aims to reverse engineer the components responsible for task completion to understand their behavior. Previous MI research has focused on linguistic tasks like Indirect Object Identification (IOI). In this paper, we investigate the ability of GPT-2 small to handle binary truth values by analyzing its behavior with syllogistic prompts, such as \lminput{Statement A is true. Statement B matches statement A. Statement B is}, which requires more complex logical reasoning compared to IOI. Through our analysis of several syllogism tasks of varying difficulty, we identify multiple circuits that mechanistically explain GPT-2’s logical reasoning capabilities and uncover binary mechanisms that facilitate task completion, including the ability to produce a negated token that does not appear in the input prompt through negative heads. Our evaluation using a faithfulness metric shows that a circuit comprising five attention heads achieves over 90\% of the original model’s performance. By relating our findings to IOI analysis, we provide new insights into the roles of certain attention heads and MLPs in LMs. We believe these insights contribute to a broader understanding of model reasoning and benefit future research in mechanistic interpretability.
\end{abstract}

\section{Introduction}
Despite the success of Large Language Models (LLMs) and their amazing capabilities, these models remain largely opaque and function as black boxes. Mechanistic interpretability has emerged as a field dedicated to mitigate this conceptual gap. By analyzing how LMs solve specific tasks \citep{wang2022interpretabilitywildcircuitindirect, hanna2023doesgpt2computegreaterthan, merullo2024circuitcomponentreusetasks}, studying emergent behaviors \citep{arditi2024refusallanguagemodelsmediated}, and identifying patterns within their architectures \citep{gurnee2024universalneuronsgpt2language}, researchers aim to unravel the inner workings of LMs. 
Even though great progress has been made, significant gaps remain in understanding LMs even on basic tasks. 

GPT-2 is a family of representative LLMs that has been frequently studied in mechanistic interpretability literature. An exemplary case is analyzing its ability to do Indirect Object Identification task~\citep{wang2022interpretabilitywildcircuitindirect}, which reverse engineers how GPT-2 correctly predicts the final token in sentences like \lminput{When Mary and John went to the shops, John gave a bottle of milk to}. Such mechanistic analysis begins with the output and traces back to identify the architectural components relevant to the task, termed as \textit{circuit}. GPT-2 small has been shown to be competent in such linguistic tasks, however its ability and mechanism to perform logic reasoning remains uncertain. Specifically, it lacks the capability to coherently answer true-false questions, such as \lminput{True or False? Dogs have four legs.}. To investigate how GPT-2 represents and processes truth values, we utilize syllogism tasks --- a classic form of logical reasoning involving premises and a conclusion. By applying similar mechanistic analysis to syllogistic prompts, we aim to discover the circuits that are relevant to the task and interpret the internal mechanisms GPT-2 uses when handling truth values. 

This paper builds on previous interpretability research by focusing on how GPT-2 handles syllogistic prompts. We use true-false syllogism tasks where truth values are assigned to premise statements and the model is prompted to predict the truth value of the conclusion. We define three prompt formats to probe binary reasoning. The \textit{Simple Syllogism} (SS) presents direct entailment, e.g., \lminput{Statement A is true. Statement B matches statement A. Statement B is}. The \textit{Opposite Syllogism} (OS) inverts this logic, requiring negation, e.g., \lminput{Statement A and Statement B are opposite. Statement A is true. Statement B is}. The \textit{Complex Syllogism} (CS) adds one or more distractor premises irrelevant to the inference, e.g., \lminput{Statement A is true. Statement B matches statement A. Statement C is false. Statement B is}, where the distractor is \lminput{Statement C is false}.

\begin{figure*}[t]
\begin{center}
\includegraphics[width=\textwidth]{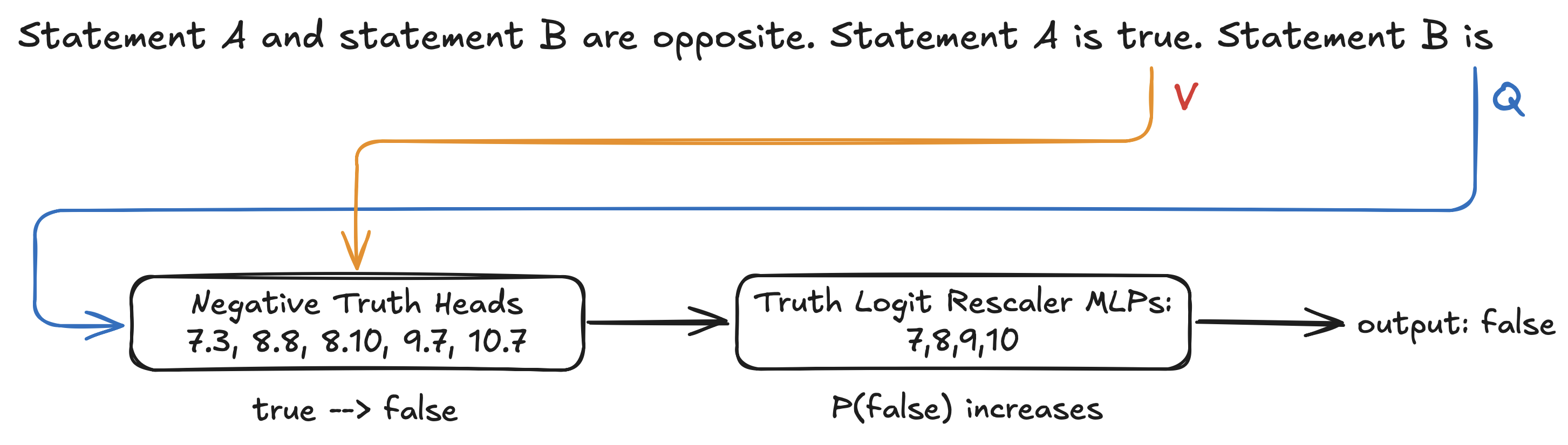}
\end{center}
\caption{Opposite Syllogism Circuit. The blue arrow represents queries, and the yellow arrow represents values. Negative Truth Heads perform negation of the truth value present in the prompt while the Truth Logit Rescaler MLPs rescale the residual stream to select the correct truth value.}
\label{figure:OS_diagram}
\end{figure*}

Our approach includes two mechanistic interpretability techniques: Path Patching and Logit Lens. Path patching \citep{wang2022interpretabilitywildcircuitindirect} determines the importance of a computational component in solving a task by replacing part of the model's forward pass with activations from a different distribution. Logit Lens \citep{Nostalgebraist} applies the model's unembedding matrix at different stages of the residual stream, exposing logits and offering insights into the function of specific components during the model's processing. Using these techniques, we apply a mechanistic lens to uncover how LMs perform complex reasoning tasks and identify the key components that drive their decisions. Specifically, we examine the internal mechanisms responsible for \negation{} and \reinforcement{} of truth values. Evaluation with a circuit faithfulness metric shows that a circuit of three attention heads can recover 90\% of the original model’s performance on SS prompts. For OS prompts, a circuit of five attention heads and four MLPs nearly recovers the performance of the full GPT-2 model, achieving roughly 85\% faithfulness. The structure of the OS circuit is shown in Figure \ref{figure:OS_diagram}.

Throughout our investigation into how GPT-2 processes syllogisms, we uncover several insights into its internal mechanisms and reasoning capabilities. Our contributions include: 

\begin{enumerate}
    \item \textbf{Discover Syllogism-Specific Circuits}: We discover circuits that represent the internal mechanisms through which GPT-2 solves syllogisms of varying complexity. 
    
    \item \textbf{Identify a Negation Mechanism}: We identify a novel mechanism for outputting the \negation{} of a truth value. Attention heads suppress the truth logit and MLPs modulate the \negation{} of the truth logit in the output distribution.
    
    \item \textbf{Explain Importance of Negative Components}: Through analysis of a pair of semantically opposite tasks (SS and OS), we demonstrate that components critical for one task often have corresponding negative counterparts that play a causally important role in the opposite task. This provides new insights into how language models process and represent binary pairs of tokens.
    
\end{enumerate}

\section{Preliminary}
\noindent\textbf{Transformers Circuits}

We provide a brief overview of GPT-2 following the notation from \citet{elhage2021mathematical}. GPT-2 is a decoder-only transformer with 12 layers; each layer contains 12 attention heads and one MLP. Input tokens \( t \) are embedded into the initial residual stream state \( x_0 \). The residual stream, a core intermediate representation, is updated additively as it passes through each layer's components

At layer \( i \), the residual stream \( x_{i-1} \) is processed by the layer’s components and updated as follows:
\begin{equation*}
\begin{split}
x_i =\ & x_{i-1} \\
       & + \text{AttentionHeads}(x_{i-1}) \\
       & + \text{MLP}\bigl(x_{i-1} + \text{AttentionHeads}(x_{i-1})\bigr).
\end{split}
\end{equation*}
Here, the attention heads process \( x_{i-1} \) in parallel, and their combined output is added back to the residual stream before passing through the MLP, whose output is then added residually to form \( x_i \).

Each attention head is parameterized by four matrices: query \( W_Q \), key \( W_K \), value \( W_V \), and output \( W_O \), which form the following composite matrices:
\[
W_{QK} := W_Q^\top W_K, \quad W_{OV} := W_O W_V.
\]
Using these matrices, along with the embedding matrix \( W_E \) and the unembedding matrix \( W_U \), the attention computation for each head decomposes into two core circuits. 

The \emph{Query-Key (QK) circuit}, defined as \( W_E^\top W_{QK} W_E \), provides the attention scores for every query–key token pair. Intuitively, each entry describes how much a given query token wants to attend to a given key token, providing insights \emph{where} information flows within the model. 

The \emph{Output-Value (OV) circuit}, defined as \( W_U W_{OV} W_E \), determines \emph{what} information is transferred to the output logits when a token is attended to.

This formulation also allows transformers to be represented as a computational graph, where nodes correspond to components like attention heads or MLPs, and edges represent learned weights. Circuits, subgraphs specialized for particular tasks, can then be identified and studied mechanistically.

\noindent\textbf{Indirect Object Identification} \citet{wang2022interpretabilitywildcircuitindirect} analyzed GPT-2 small’s performance on the IOI task, where the model must predict the indirect object (IO) in sentences like: \lminput{When Mary and John went to the store, John gave a bottle of milk to Mary.} The correct prediction is \lminput{Mary}, not the repeated subject \lminput{John}.

A human-interpretable strategy to solve IOI involves three steps: (1) identify all names in the sentence, (2) remove duplicates, and (3) output the remaining name. GPT-2 small mirrors this algorithm through three distinct attention head groups: Duplicate Token Heads detect repeated names, attending from the second mention back to the first; S-Inhibition Heads suppress repeated tokens; and Name Mover Heads copy the correct IO into the output via attention.

\noindent\textbf{Path Patching}  
Path patching is an intervention-based interpretability method for circuit discovery \citep{wang2022interpretabilitywildcircuitindirect}. It utilizes two prompt distributions: the original task distribution \( p_{\text{orig}} \), and a corrupted distribution \( p_{\text{new}} \) designed to break task-relevant behavior. First, the model is run on both distributions and each computational node's activations are cached. Then, a forward pass is performed on \( p_{\text{orig}} \) where the activation at a specific node \( F \) (e.g., an attention head) is replaced with its counterpart from \( p_{\text{new}} \), while the rest of the model remains unchanged. Next, the resulting activation at a downstream node \( G \) is cached and patched into a forward pass on \( p_{\text{new}} \). The causal impact of the path \( F \rightarrow G \) is quantified by measuring the change in logit difference. A substantial drop indicates that \( F \) is causally important for the model’s behavior on the task.

\noindent\textbf{Logit Lens}  
The logit lens \citep{Nostalgebraist} is an interpretability method that projects the hidden state of a computational node, $h$, into the model’s token space. It applies layer normalization followed by the unembedding matrix:
\[
\text{LogitLens}(h) = \text{LayerNorm}(h) W_U.
\]
This yields a distribution over tokens, revealing which outputs the model would favor if it predicted directly from that point.


\section{Understand How GPT-2 Process Syllogisms}
Syllogisms \citep{aristotle_prior_analytics} offer an effective way to analyze a LM's reasoning capacity. Rather than analyzing a broad range of facts in a syllogistic format, we narrow our focus to a simpler set of propositions and declarative statements such as: \lminput{Statement A is true. Statement B matches statement A. Statement B is true.} We define three types of syllogisms: \textbf{Simple Syllogism (SS)}, \textbf{Opposite Syllogism (OS)}, and \textbf{Complex Syllogism (CS)}. A complete example of each type is provided in Table \ref{table:syllogism_examples}. We define the logit difference for the syllogism task family as follows.  Let the answer set be $S = \{\text{true}, \text{false}\}$ with the correct answer $x \in S$, and the incorrect answer $\neg x  \in S$. The logit difference (LD) is then given by:  
\[
\text{LD} = \text{logit}(x) - \text{logit}(\neg x).
\]
A positive logit difference indicates that the first logit is more probable, while a negative logit difference suggests the second logit is more probable. $e^{LD}$ represents how many times more likely the model will predict $x$ compared to $\neg x$. Thus, for the SS format, GPT-2 small is 6.4077 times more likely to predict the correct truth value.

To quantify how well a circuit preserves model behavior, we use the \textbf{faithfulness} metric. Let \( ALD(\mathcal{M}) \) denote the average logit difference (ALD) of the full model \( \mathcal{M} \), and \( ALD(\mathcal{C}) \) that of a circuit \( \mathcal{C} \). The faithfulness metric is defined as:
\[
\text{Faithfulness} = \left| ALD(\mathcal{M}) - ALD(\mathcal{C}) \right|.
\]
A lower value indicates that the circuit faithfully recovers the model’s behavior on the task.

\renewcommand{\arraystretch}{1.5} 
\begin{table*}[t]
    \centering
    \begin{tabularx}{\textwidth}{lXc}
    \toprule
    \textbf{Type} & \textbf{Example Syllogism} & \textbf{Avg. Logit Diff.}\\
    \midrule
    Simple & \textit{Statement A is true. Statement B matches statement A. Statement B is} \textcolor{red}{\textit{true}} & 1.8575 \\
    Opposite & \textit{Statement A and statement B are opposites. Statement A is true. Statement B is}  \textcolor{red}{\textit{false}} & 1.2123  \\
    Complex & \textit{Statement A is true. Statement B matches statement A. Statement C is false. Statement B is} \textcolor{red}{\textit{true}} & 1.3105 \\ \bottomrule
    \end{tabularx}
    \caption{Examples of syllogism types with their corresponding average logit differences over datasets of 500 prompts. The LM is expected to predict the red tokens. We create these distributions of syllogism by replacing letters and truth values.}
    \vskip -0.15in
    \label{table:syllogism_examples}
\end{table*}

\subsection{Simple Syllogism}
We frame the SS task with the following human-interpretable algorithm: (1) Identify the single truth value token in the prompt; (2) Output the truth value token. Construction of the SS dataset can be found in Appendix \ref{syllogism_dataset_construction}.

\begin{figure*}[t]
\centering
\subfloat[MLP Direct Effect on Logits\label{figure:ss_mlp_direct_effect_w_attn_head}]{
\includegraphics[trim=20 10 20 10, clip,width=0.30\textwidth]{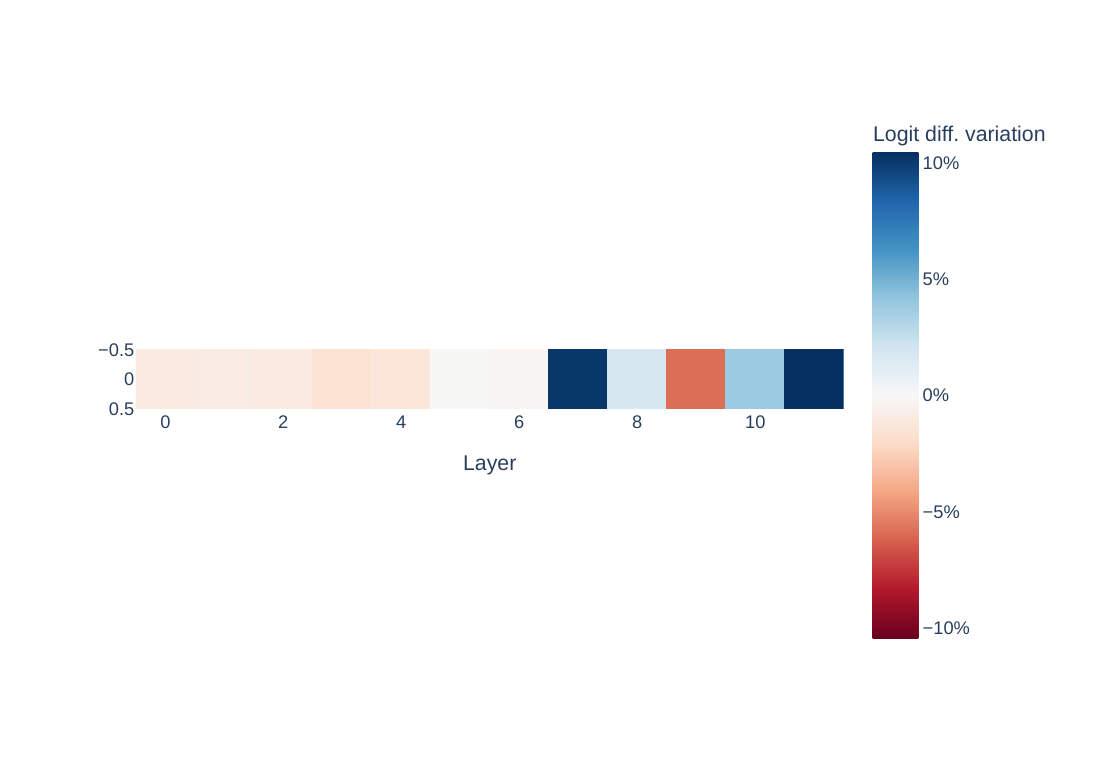}
}
\hfill
\subfloat[Attention Head Direct Effect on Logits\label{figure:ss_attn_direct_effect}]{
\includegraphics[trim=20 10 20 10, clip, width=0.30\textwidth]{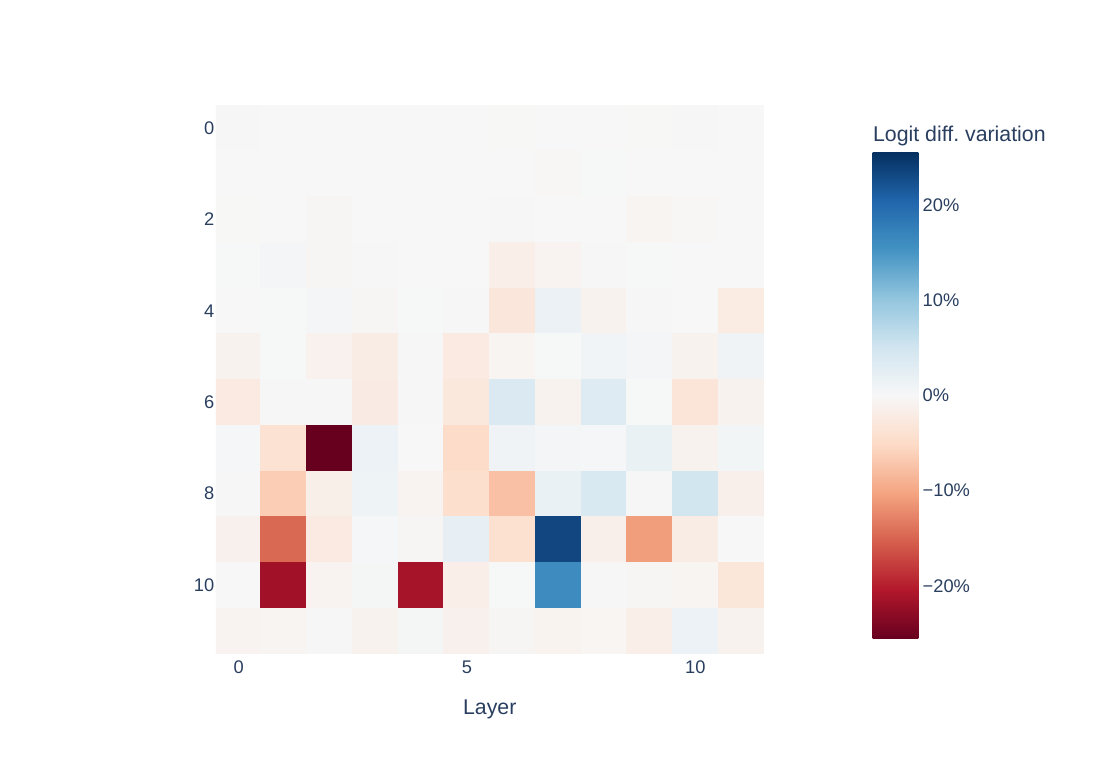}
}
\hfill
\subfloat[QK Circuit Visualization (Head 7.2)\label{figure:ss_7_2_qk}]{
\includegraphics[trim=20 10 20 10, clip,width=0.30\textwidth]{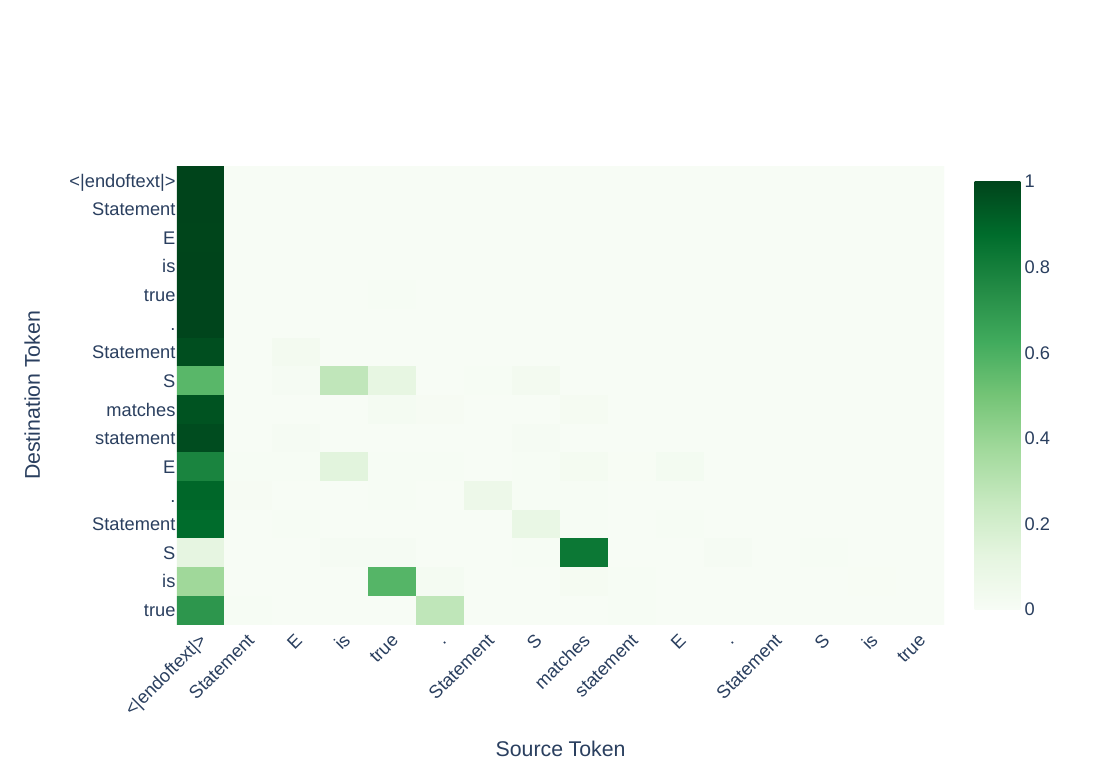}
}
\caption{Path Patching and QK Analysis on Simple Syllogism Prompts.}
\label{figure:Simple_Syllogism_Path_Patching_Logits}
\end{figure*}

\noindent\textbf{Truth Heads}
We begin by applying path patching to determine which attention heads and MLPs influence the model’s output logits on SS prompts. As shown in Figure~\ref{figure:ss_mlp_direct_effect_w_attn_head}, MLP layers have minimal direct effect on the logits, suggesting they are not essential for solving the SS task. We explore this further in Appendix~\ref{do_mlps_matter_ss}.

In contrast, Figure~\ref{figure:ss_attn_direct_effect} reveals that several attention heads in the later layers, particularly heads 7.2, 9.1, 9.9, 10.1, and 10.4, contribute substantially to logit differences. To understand the behavior of these heads, we analyze their attention patterns using their $QK$ circuits. Specifically, for each attention head \( h \), we compute the raw attention score:

\[
A^h =  t^\top W_E^\top W_{QK}^h W_E t,
\]

which captures how much each query token attends to each key token in the vocabulary space. We find that these heads exhibit similar induction-like attention patterns: they predominantly attend to the final token corresponding to the truth value. 

We provide a visualization of the most influential head, 7.2, in Figure~\ref{figure:ss_7_2_qk}, along with the top $K = 3$ query--key token pairs in Table~\ref{table:qk_top_pairs_ss}. For reference, we use the example SS prompt:
\lminput{Statement E is true. Statement S matches statement E. Statement S is true}. Across top heads we consistently observe two high-scoring token pairs: \((\texttt{S}, \texttt{matches})\) and \((\texttt{is}, \texttt{true})\). The first pair indicates that GPT-2 has developed a logical understanding of equivalence between the two statements—effectively computing \( matches(S, E) \)—while the second pair shows it retrieving the correct truth value based on this relationship. This consistent behavior leads us to call these attention heads \textit{Truth Heads}.

\renewcommand{\arraystretch}{1.5}
\begin{table*}[t]
    \centering
    \begin{tabularx}{\textwidth}{lXXX}
    \toprule
    \textbf{Head} & \textbf{1st Highest Q--K Pair} & \textbf{2nd Highest} & \textbf{3rd Highest} \\
    \midrule
    7.2  & \textbf{0.8304:} \mbox{\texttt{['S', 'matches']}} & \textbf{0.5749:} \mbox{\texttt{['is', 'true']}} & 0.2750: \mbox{\texttt{['true', '.']}} \\
    10.1 & \textbf{0.7139:} \mbox{\texttt{['is', 'true']}} & \textbf{0.5524:} \mbox{\texttt{['S', 'matches']}} & 0.4258: \mbox{\texttt{['S', 'true']}} \\
    10.4 & \textbf{0.6833:} \mbox{\texttt{['is', 'true']}} & \textbf{0.6637:} \mbox{\texttt{['S', 'matches']}} & 0.5063: \mbox{\texttt{['.', 'Statement']}} \\
    \bottomrule
    \end{tabularx}
    \caption{
    Top 3 highest-scoring query--key token pairs from the attention pattern scores of the most influential heads in the SS format.
    }
    \label{table:qk_top_pairs_ss}
\end{table*}

To test whether Truth Heads depend on earlier attention heads, we repeat path patching on their query, key, and value inputs. We find that earlier heads have minimal effect, suggesting that the Truth Heads operate independently. To verify their sufficiency we build a minimal circuit $C_{SS}$ consisting only of the Truth Heads. $C_{SS}$ faithfully recovers the predictions of the model, achieving an average logit difference of 1.9286, effectively matching the performance of the GPT-2 small on the task. The Truth Heads' $QK$ circuit consistently directs attention to the correct truth value earlier in the prompt while their $OV$ circuit copies that value into the residual stream at the final token. Using the logit lens on truth heads confirm their output strongly favors the correct truth value.

\renewcommand{\arraystretch}{1.4}

\begin{table*}[t]
\centering
\begin{tabularx}{\textwidth}{X}
\toprule
\textbf{Top QK Pairs (Head 10.7)} \\
\midrule
\textbf{0.892:} (‘is’, \textbf{‘true’}), 
\textbf{0.772:} (‘statement’, ‘E’), 
\textbf{0.685:} (‘Statement’, ‘S’), 
\textbf{0.662:} (‘Statement’, ‘S’), 
\textbf{0.459:} (‘is’, ‘oppos’) \\
\bottomrule
\end{tabularx}


\centering
\begin{tabularx}{\textwidth}{lX>{\raggedright\arraybackslash}X}
\toprule
\textbf{Stage} & \textbf{Top Logits} & \textbf{Bottom Logits} \\
\midrule
After OV from Head 10.7 &
depot, rink, carp, Dj, Hack, DJ, Gaz, Phillips, District, TTC &
\textbf{‘true’}, ‘True’, ‘TRUE’, \textbf{‘true’}, ‘untrue’, ‘Null’ \\
\midrule
After MLP Layer 10 &
\textbf{‘true’}, \textcolor{red}{\textbf{‘false’}}, \textbf{‘True’}, \textcolor{red}{\textbf{‘False’}}, infinite, truly &
blitz, ombo, plateau, corrid, tradem, emale, Citiz, sugg \\
\bottomrule
\end{tabularx}
\caption{Top QK pairs in Head 10.7 strongly attend to truth-related tokens (e.g., ‘is’, ‘true’). Initially, the OV output does not rank truth tokens highly. However, after the MLP layer, both \textbf{‘true’} and \textcolor{red}{\textbf{‘false’}} become top-ranked, indicating the MLP can help produce the opposite token, even when it is not in the prompt}
\label{table:mlpflip}
\end{table*}

\noindent\textbf{Negative Heads in Simple Syllogism}  
In addition to the Truth Heads, we identify a distinct group of heads—such as 9.7, 10.7, and 11.10—that exhibit attention patterns similar to the Truth Heads but are not essential for solving the task. Notably, mean-ablating head 10.7 improves model performance beyond the baseline. Head 10.7 has previously been characterized as a negative head in prior work \citep{wang2022interpretabilitywildcircuitindirect} where it was shown to reduce the logit of specific output tokens. We hypothesize these negative heads encode the logit of the incorrect class in a binary setting. This aligns with findings from the copy suppression literature \citep{mcdougall2023copysuppressioncomprehensivelyunderstanding} where head 10.7 was also found to suppress certain tokens. To test this, we turn to the Opposite Syllogism format.

\subsection{Opposite Syllogism}
To test our hypothesis surrounding negative heads, we investigate the model’s behavior on \emph{opposite syllogisms} (OS). We define a human-interpretable algorithm for this task in three steps: (1) Identify the single truth value token in the prompt, (2) Negate the truth value token, and (3) Output the negated token. Details on dataset construction are provided in Appendix~\ref{syllogism_dataset_construction}. For reference, we use the example OS prompt:
\lminput{Statement E and statement S are opposites. Statement E is true. Statement S is false}.

\begin{figure*}[h]
    \centering
    \subfloat[MLP Direct Effect on Logits\label{figure:os_mlp_direct_effect_w_attn_head}]{
        \includegraphics[trim=20 10 20 10, clip,width=0.30\textwidth]{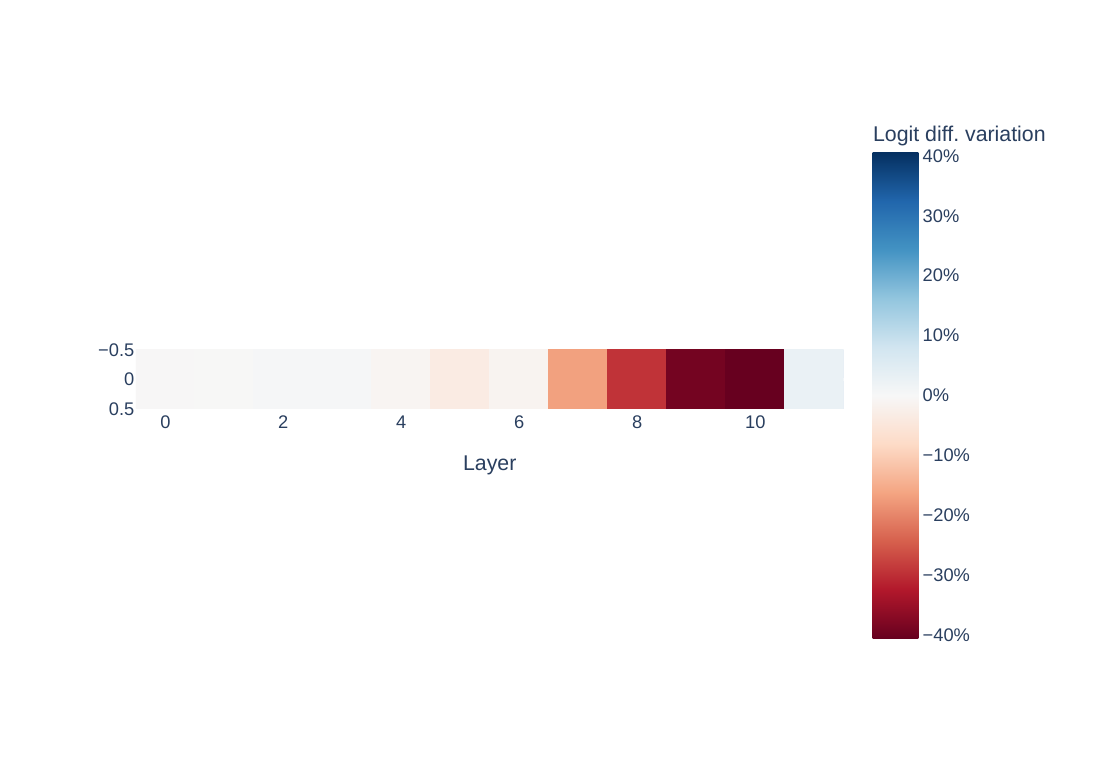}
    }
    \hfill
    \subfloat[Attention Head Direct Effect on Logits\label{figure:os_attn_direct_effect}]{
        \includegraphics[trim=20 10 20 10, clip, width=0.30\textwidth]{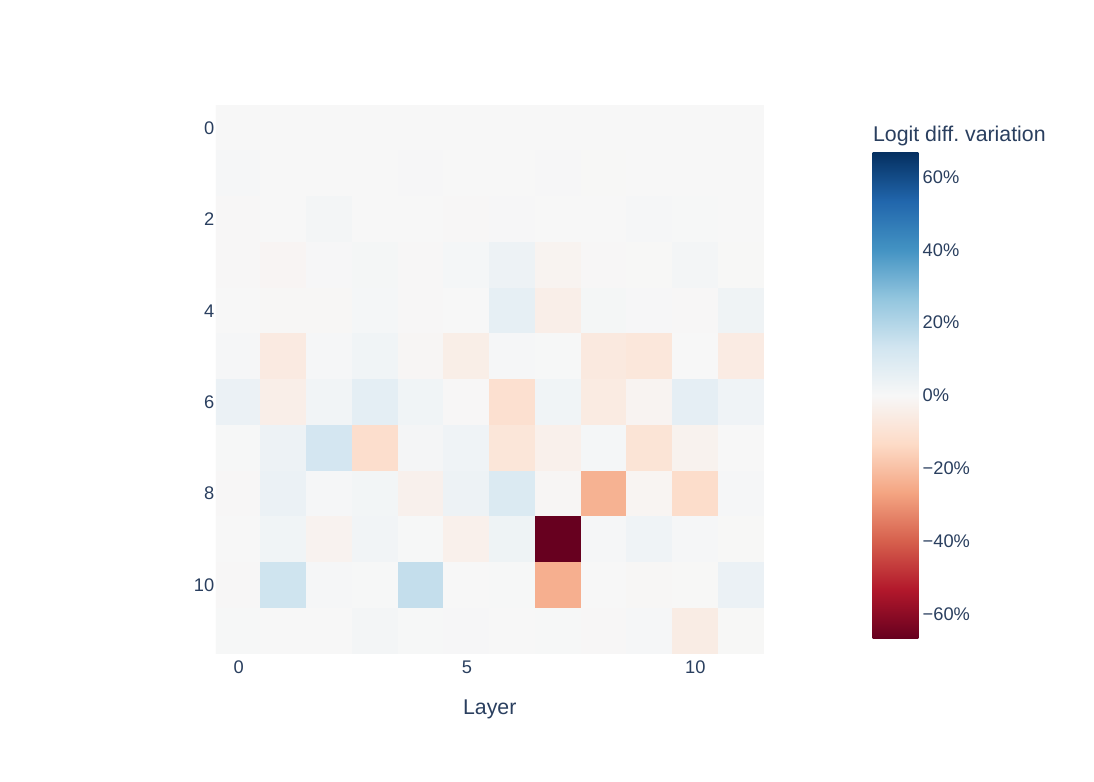}
    }
    \hfill
    \subfloat[QK Circuit Visualization (Head 10.7)\label{figure:os_10_7_qk}]{
        \includegraphics[trim=20 10 20 10, clip,width=0.30\textwidth]{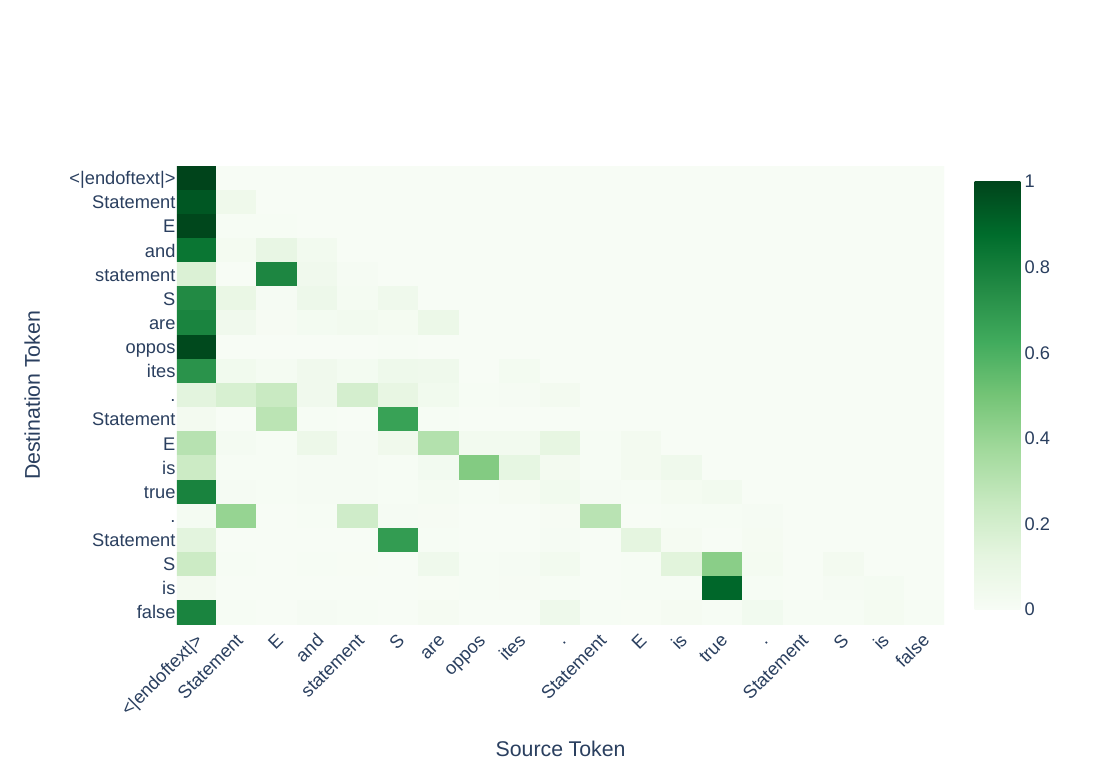}
    }
    \caption{Path Patching and QK Analysis on Opposite Syllogism Prompts.}
    \label{figure:Opposite_Syllogism_Path_Patching_Logits}
\end{figure*}

\noindent\textbf{Negative Truth Heads.}
We begin by identifying components that directly influence the model’s output on OS prompts.  
Path patching shows that ablating attention heads 7.3, 8.8, 8.10, 9.7, and 10.7 leads to a significant drop in logit difference (Figure~\ref{figure:os_attn_direct_effect}).  
To understand their role, we analyze each head’s output by applying its $OV$ matrix to the MLP-extended embedding basis, following prior techniques from \citet{wang2022interpretabilitywildcircuitindirect} and \citet{mcdougall2023copysuppressioncomprehensivelyunderstanding}:
\[
W_U \, W_{OV}^{h}\,\mathrm{MLP}_0\!\bigl(W_E\bigr).
\]
These heads consistently attend to the truth value token in the prompt (e.g., \texttt{is}, \texttt{true})—mirroring behavior observed in the SS setting.  
However, their influence on the logits differs. Some heads, like 8.8 and 5.1, promote the truth token to the top logits and function as standard \emph{Truth Heads}.  
In contrast, heads such as 7.3, 8.10, 9.7, 10.7, and 11.10 suppress the truth token into the bottom logits (Table~\ref{table:mlpflip}).  
We refer to this group as \emph{Negative Truth Heads}.

Unlike the SS format, MLPs corresponding to these heads are crucial: ablating them significantly reduces performance (Figure~\ref{figure:os_mlp_direct_effect_w_attn_head}).  
Furthermore, path patching the queries of these Negative Truth Heads confirms that they operate independently, with no significant upstream influence, mirroring earlier findings from SS where only components with direct logit impact matter.

\medskip
\noindent\textbf{Mechanistic Interpretation}
The $QK$ circuit of each Negative Truth Head reliably identifies the truth token—fulfilling Step 1 of the OS algorithm.  
However, their $OV$ projection suppresses this token into the bottom logits.  
The associated MLP then rescales the residual stream to promote the opposite of the suppressed token into the top logits, completing Steps 2 and 3.  
We refer to these components as \emph{Truth Logit Rescaler MLPs}.

This attention–MLP sequence forms a mechanistic pathway for learned negation: the model suppresses a truth token and then elevates its negation for output.  
Table~\ref{table:mlpflip} captures the entire trajectory of such a token: from attention-induced suppression to MLP-driven recovery.  
This reveals how the model predicts a correct token not seen in the prompt, using the suppression of the incorrect token as a signal for its opposite. While this attention–MLP mechanism effectively negates the truth value in opposite syllogisms, we observe a consistent asymmetry: the negation process is more reliably triggered when the input token is \texttt{true}, resulting in \texttt{false} predictions. In contrast, when the prompt contains \texttt{false}, the model often retains \texttt{false} as the dominant logit rather than flipping to \texttt{true}.

\medskip
\noindent\textbf{Circuit Faithfulness}
To test sufficiency, we construct a circuit $C_{\text{OS}}$ using only the Negative Truth Heads and their associated MLPs.  
This circuit recovers approximately $85\%$ of GPT-2 small’s performance, demonstrating it is a faithful subcircuit for solving the OS task.  
A schematic of this circuit is shown in Figure~\ref{figure:OS_diagram}.

\noindent\textbf{Reversal of Head Behavior Between Tasks}
Interestingly, the same heads that negatively affected logit difference in the SS task—like 10.7—now play a constructive role in OS.  
This reversal demonstrates that the model reuses certain components in complementary tasks where their function flips to support inverse outcomes.  
We extend this finding in Appendix~\ref{generality_of_binary_task}, where we test circuit generalization and transferability across other binary pairs beyond \texttt{true}/\texttt{false}, further supporting the idea that GPT-2 small represents logical negation via attention and MLPs.



\subsection{Complex Syllogism}

The complex syllogism task expands on the previous setups by introducing a misleading, redundant statement. The objective is to determine whether GPT-2 Small can still arrive at the correct conclusion in the presence of potentially confusing information. 

We begin with path patching to identify which attention heads and MLPs directly influence the model’s output logits. The results are similar to the path patching results of the OS format. More specifically, MLPs in layers 8,9, and 10 positively influence the logit difference. Similar heads—specifically, heads 7.3, 8.8, 8.10, 9.7, and 10.7— were found to positively influence logit difference but with different behavior. Heads 9.7, 10.7, and 11.10 attend to both truth tokens in the prompt but place greater emphasis on the incorrect (redundant) truth token. Inspecting their logits, we find these heads continue to perform a suppression operation as observed previously, and we therefore classify them as \textit{Negative Truth Heads} in the CS format and do not investigate them further. In contrast, heads 7.3, 8.8, and 8.10 exclusively attend to the incorrect truth value, and we study them more closely to understand their contribution. We refer to these as \textit{Truth Modulation Heads}.

To interpret the behavior of the Truth Modulation Heads, we analyze how their outputs project onto the unembedding vectors of the truth tokens. Let \( W_U[\texttt{true}] \) and \( W_U[\texttt{false}] \) represent the unembedding directions for the correct and incorrect truth tokens respectively. For a head output \( h_i(X) \), we compute the logit contribution to token \( t \in \{\texttt{true}, \texttt{false}\} \) as
\[
\langle h_i(X), W_U[t] \rangle.
\]
This dot product reflects how strongly head \( h_i \) pushes the residual stream toward generating token \( t \). We scatter plot each head’s attention probability on the truth value token against the corresponding logit contribution along the true direction. Two distinct patterns emerge across the heads. First, the correct token is consistently ignored by these heads, receiving very low attention probability. Second, the incorrect token shows two opposing trends depending on the head: in some cases, the attention probability and logit contribution are positively correlated, suggesting that the head reinforces the incorrect truth value. We refer to such heads as \textit{Correct Truth Inhibition Heads}. In other cases, the relationship is negative—the more attention the head gives to the incorrect token, the more it pushes away from the incorrect truth direction. This effectively reinforces the correct token and we refer to these as \textit{Correct Truth Reinforcement Heads}.

These behaviors are further supported by examining the top and bottom logits. In inhibition heads, the incorrect token consistently appears among the top logits while in reinforcement heads it appears among the bottom logits. This supports the interpretation that Truth Modulation Heads implement a binary operation: either reinforcing or inhibiting the direction of the incorrect token, which indirectly determines the correct output.

We then investigate whether the Negative Truth Heads and Truth Modulation Heads influence one another. Path patching reveals that neither group affects the other directly, although both are influenced by similar upstream sources. Moreover, ablating one group does not destroy faithfulness, confirming that the groups can operate independently. This redundancy is consistent with the findings of \citet{mcgrath2023hydraeffectemergentselfrepair}, who describe the Hydra Effect in language models, where multiple pathways can implement the same behavior.

Path patching reveals that both the Negative Truth Heads and the Truth Modulation Heads receive input from a shared set of upstream heads, primarily located in Layers 0 and 5. In Layer 5, heads 5.1 and 5.5 exhibit classic induction patterns as described by \citet{elhage2021mathematical}, attending from the conclusion line (e.g., \lminput{Statement B is}) back to a matching premise. This effectively links the conclusion to its logical source. In Layer 0, other heads attend to repeated statement identifiers (e.g., \lminput{Statement B}) and influence both the key vectors of downstream heads and the values used by the Induction Heads. These heads appear to detect repeated statements and function as \textit{Duplicate Statement Identifier Heads}, marking the reuse of information—an essential step in a natural deduction process.

At inference time, the Induction Heads serve as a routing mechanism: they either connect the conclusion to the premise via the \lminput{matches} relation or directly extract the incorrect truth value from the conclusion line. This information is then processed by the Negative Truth Heads or Truth Modulation Heads to generate the final token.


\section{Discussion}



\noindent\textbf{Connection to IOI}
Despite overlap with the Name Mover Heads from the IOI task, we find (Negative) Truth Heads reflect broader functionality, particularly negate rather than simply copy. From an IOI perspective, the Negative Truth Heads were initially interpreted as negative copy heads due to their tendency to replicate the tokens they attend to. However, in the opposite syllogism task, the correct answer is not explicitly present in the prompt. Consequently, these heads cannot simply copy the attended truth value to produce the correct answer. This provides strong evidence that Negative Truth Heads encode the direction of the less contextualized logit in a binary setting, effectively operating in the antidirection. We believe this behavior remained unnoticed in IOI because, in that context, $\text{Mary} \neq \neg \text{John}$. Similarly, many of the Truth Modulation Heads align with the S-Inhibition category from IOI, suggesting a shared functional role. We identify the Correct Truth Inhibition Heads as the original inhibition heads from IOI, given their role in reinforcing focus on the incorrect token. This expanded understanding highlights how heads previously characterized in IOI tasks can exhibit more nuanced and adaptable behaviors in different contexts.

\begin{figure*}[t]
    \centering
    \subfloat[Correct Truth Reinforcement Head 8.10 Projection along truth embedding versus attention probability on truth value\label{figure:refined_reinforcment}]{\includegraphics[width=0.45\textwidth]{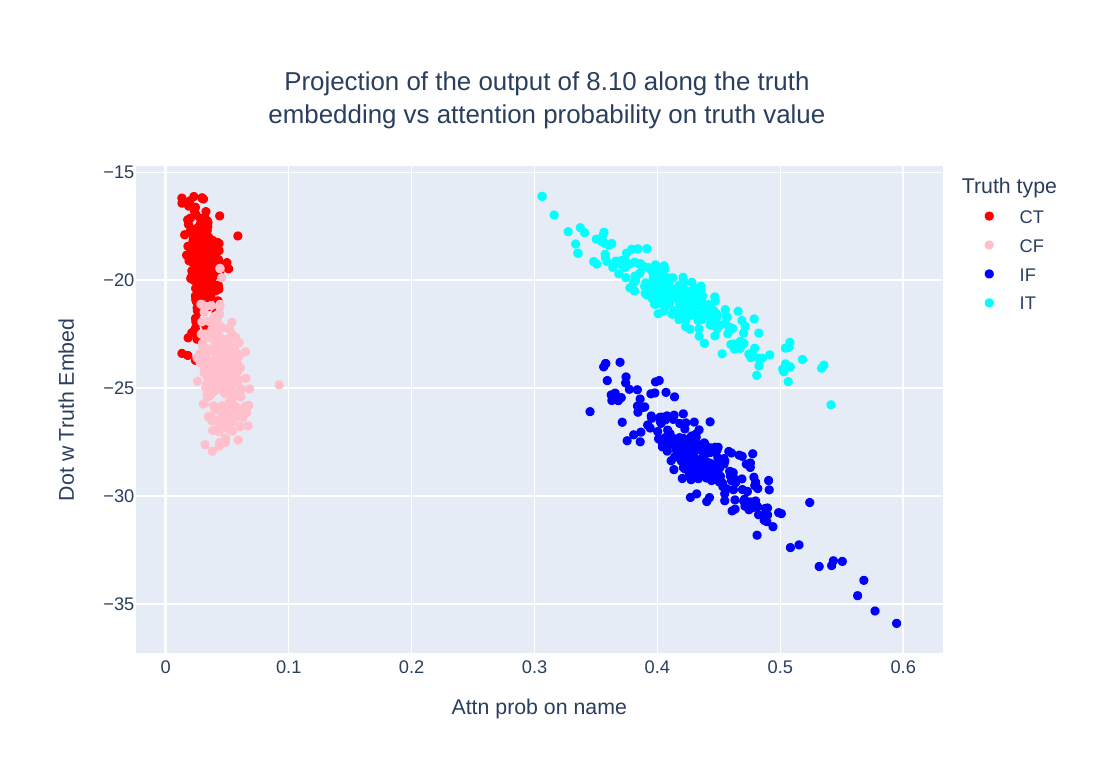}}
    \hfill
    \subfloat[Correct Truth Inhibition Head 8.8 Projection along truth embedding versus attention probability on truth value \label{figure:refined_inhibition}]{\includegraphics[width=0.45\textwidth]{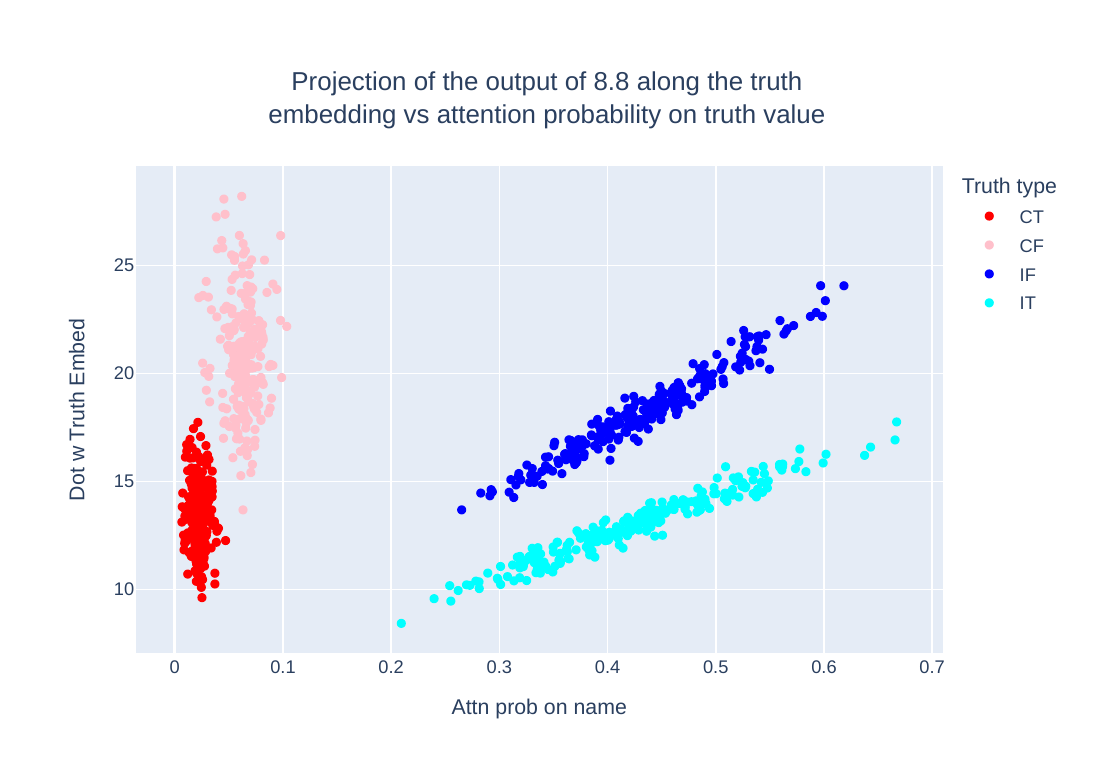}}
    \caption{The Truth Modulation group with refined truth types}
    \vskip -0.15in
    \label{figure:refined_truth_modulation_heads}
\end{figure*}

\noindent\textbf{Clustering of Truth Modulation Heads}
We observe distinct clusters in both groups of Modulation Heads. To refine our truth types, we categorize truth values into four types: correct true (CT), correct false (CF), incorrect true (IT), and incorrect false (IF). This results in two natural pairings: (CT, IF) and (CF, IT). As shown in Figure~\ref{figure:refined_reinforcment} and \ref{figure:refined_inhibition}, \texttt{false} (IF or CF) has larger projections on the truth embedding. We believe that this asymmetry not only reflects the internal bias of truth values learned from the training corpus, but also resembles the behavior observed in the Opposite Syllogism task, where negation was easier with \texttt{true}. Although we do not rigorously analyze this connection, it may reflect a broader model bias toward negative truth values or a negation-like structure in its internal representations.

\noindent\textbf{Scalability of Results}
We extend our simple and opposite syllogism formats to larger models such as GPT-2 XL, Pythia 1.4B, Qwen3-1.7B, and LLaMA3.2-1B. (See Appendix \ref{section:larger_models}). 

\section{Related Works}
\noindent\textbf{Mechanistic Interpretability} Mechanistic Interpretability research offers various techniques to reverse-engineer model behavior and identify important components responsible for a model's performance. In addition to Path Patching \citep{wang2022interpretabilitywildcircuitindirect}, there are other patching methods including Attribution Patching \citep{nandaattributionpatching}, causal mediation analysis \citep{meng2023locatingeditingfactualassociations, 10.1145/3501714.3501736, NEURIPS2020_92650b2e}, and AtP* \citep{kramár2024atpefficientscalablemethod}. Sparse Autoencoders (SAEs) have become increasingly popular for interpreting features \citep{bricken2023monosemanticity, marks2025sparsefeaturecircuitsdiscovering}. Earlier works such as Neuron Shapley \citep{ghorbani2020neuronshapleydiscoveringresponsible} introduce a framework that quantifies each neuron's contribution to a deep network's performance by considering interactions among neurons. Other earlier works such as \citet{cao2021lowcomplexityprobingfindingsubnetworks, csordás2021neuralnetsmodularinspecting} employ subnetworks to investigate what model internals are needed to perform a task through probing and masking.

\noindent\textbf{Circuit Discovery in GPT-2} IOI has inspired many other circuit analysis works. \cite{hanna2023doesgpt2computegreaterthan} identify a circuit that explains GPT2's ability to predict correct year tokens when prompted with task like ``The war lasted from the year 1732 to 17''. \cite{merullo2024circuitcomponentreusetasks} rediscover the IOI circuit in GPT2-Medium and show that much of the circuit can be reused to solve the Colored Object task introduced by \cite{srivastava2023imitationgamequantifyingextrapolating}. 
\cite{nainani2024adaptivecircuitbehaviorgeneralization} explore IOI's generality by extending the prompt to include more instances of the indirect object. \citet{conmy2023automatedcircuitdiscoverymechanistic} generalize and automate the IOI-style analysis within GPT-2 small, ultimately recovering many already discovered circuits.

\noindent\textbf{Syllogisms for Assessing LLMs} 
Recent studies have explored assessing LLMs with syllogisms. \citet{eisape2024systematiccomparisonsyllogisticreasoning, ando2023evaluatinglargelanguagemodels} provide a comparative analysis on how humans and LLMs perform syllogistic reasoning. \citet{kim2025mechanisticinterpretationsyllogisticreasoning} conduct a mechanistic analysis of standard syllogisms. In contrast, our work explores syllogisms with assigned truth values, offering a distinct perspective. Furthermore, we provide novel insights into the role of MLPs in facilitating syllogistic reasoning and handling logical negation.

\section{Conclusion}
In this work, we reverse-engineered GPT-2 for three syllogism tasks of varying complexity, uncovering key insights into how GPT-2 handles binary truth values within logical tasks. In the simplest case, high faithfulness was achieved with just Truth Heads, highlighting the model's ability to maintain correct truth values with minimal components. In the opposite syllogism case, the inclusion of Negative Truth Heads and MLPs allowed the model to properly negate the truth value, demonstrating the novel negation mechanism in handling binary outcomes. In the complex case, while negation remained a key mechanism, additional heads were needed to identify and process the correct truth value to negate, reflecting the increased complexity of the task. Our findings reveal significant overlap with the IOI circuit, expanding our understanding of these computational nodes’ capabilities; however, this understanding remains limited, underscoring the need for continued interpretability research to ensure that, as such models become increasingly integrated into daily life, their logical deductions can be relied upon and their behavior held accountable.

\bibliographystyle{unsrtnat}
\bibliography{colm2025_conference}

\begin{thebibliography}{26}
\providecommand{\natexlab}[1]{#1}
\providecommand{\url}[1]{\texttt{#1}}
\expandafter\ifx\csname urlstyle\endcsname\relax
  \providecommand{\doi}[1]{doi: #1}\else
  \providecommand{\doi}{doi: \begingroup \urlstyle{rm}\Url}\fi

\bibitem[Wang et~al.(2022)Wang, Variengien, Conmy, Shlegeris, and Steinhardt]{wang2022interpretabilitywildcircuitindirect}
Kevin Wang, Alexandre Variengien, Arthur Conmy, Buck Shlegeris, and Jacob Steinhardt.
\newblock Interpretability in the wild: a circuit for indirect object identification in gpt-2 small, 2022.
\newblock URL \url{https://arxiv.org/abs/2211.00593}.

\bibitem[Hanna et~al.(2023)Hanna, Liu, and Variengien]{hanna2023doesgpt2computegreaterthan}
Michael Hanna, Ollie Liu, and Alexandre Variengien.
\newblock How does gpt-2 compute greater-than?: Interpreting mathematical abilities in a pre-trained language model, 2023.
\newblock URL \url{https://arxiv.org/abs/2305.00586}.

\bibitem[Merullo et~al.(2024)Merullo, Eickhoff, and Pavlick]{merullo2024circuitcomponentreusetasks}
Jack Merullo, Carsten Eickhoff, and Ellie Pavlick.
\newblock Circuit component reuse across tasks in transformer language models, 2024.
\newblock URL \url{https://arxiv.org/abs/2310.08744}.

\bibitem[Arditi et~al.(2024)Arditi, Obeso, Syed, Paleka, Panickssery, Gurnee, and Nanda]{arditi2024refusallanguagemodelsmediated}
Andy Arditi, Oscar Obeso, Aaquib Syed, Daniel Paleka, Nina Panickssery, Wes Gurnee, and Neel Nanda.
\newblock Refusal in language models is mediated by a single direction, 2024.
\newblock URL \url{https://arxiv.org/abs/2406.11717}.

\bibitem[Gurnee et~al.(2024)Gurnee, Horsley, Guo, Kheirkhah, Sun, Hathaway, Nanda, and Bertsimas]{gurnee2024universalneuronsgpt2language}
Wes Gurnee, Theo Horsley, Zifan~Carl Guo, Tara~Rezaei Kheirkhah, Qinyi Sun, Will Hathaway, Neel Nanda, and Dimitris Bertsimas.
\newblock Universal neurons in gpt2 language models, 2024.
\newblock URL \url{https://arxiv.org/abs/2401.12181}.

\bibitem[Nostalgebraist(2020)]{Nostalgebraist}
Nostalgebraist.
\newblock Interpreting gpt: The logit lens, 2020.
\newblock URL \url{https://www.lesswrong.com/posts/AcKRB8wDpdaN6v6ru/interpreting-gpt-the-logit-lens}.

\bibitem[Elhage et~al.(2021)Elhage, Nanda, Olsson, Henighan, Joseph, Mann, Askell, Bai, Chen, Conerly, DasSarma, Drain, Ganguli, Hatfield-Dodds, Hernandez, Jones, Kernion, Lovitt, Ndousse, Amodei, Brown, Clark, Kaplan, McCandlish, and Olah]{elhage2021mathematical}
Nelson Elhage, Neel Nanda, Catherine Olsson, Tom Henighan, Nicholas Joseph, Ben Mann, Amanda Askell, Yuntao Bai, Anna Chen, Tom Conerly, Nova DasSarma, Dawn Drain, Deep Ganguli, Zac Hatfield-Dodds, Danny Hernandez, Andy Jones, Jackson Kernion, Liane Lovitt, Kamal Ndousse, Dario Amodei, Tom Brown, Jack Clark, Jared Kaplan, Sam McCandlish, and Chris Olah.
\newblock A mathematical framework for transformer circuits.
\newblock \emph{Transformer Circuits Thread}, 2021.
\newblock https://transformer-circuits.pub/2021/framework/index.html.

\bibitem[Aristotle(c. 350 BC)]{aristotle_prior_analytics}
Aristotle.
\newblock \emph{Prior Analytics}.
\newblock Oxford University Press, Oxford, UK, c. 350 BC.

\bibitem[McDougall et~al.(2023)McDougall, Conmy, Rushing, McGrath, and Nanda]{mcdougall2023copysuppressioncomprehensivelyunderstanding}
Callum McDougall, Arthur Conmy, Cody Rushing, Thomas McGrath, and Neel Nanda.
\newblock Copy suppression: Comprehensively understanding an attention head, 2023.
\newblock URL \url{https://arxiv.org/abs/2310.04625}.

\bibitem[McGrath et~al.(2023)McGrath, Rahtz, Kramar, Mikulik, and Legg]{mcgrath2023hydraeffectemergentselfrepair}
Thomas McGrath, Matthew Rahtz, Janos Kramar, Vladimir Mikulik, and Shane Legg.
\newblock The hydra effect: Emergent self-repair in language model computations, 2023.
\newblock URL \url{https://arxiv.org/abs/2307.15771}.

\bibitem[Nanda(2023)]{nandaattributionpatching}
Neel Nanda.
\newblock Attribution patching: Activation patching at industrial scale, 2023.
\newblock URL \url{https://www.neelnanda.io/mechanistic-interpretability/attribution-patching}.

\bibitem[Meng et~al.(2023)Meng, Bau, Andonian, and Belinkov]{meng2023locatingeditingfactualassociations}
Kevin Meng, David Bau, Alex Andonian, and Yonatan Belinkov.
\newblock Locating and editing factual associations in gpt, 2023.
\newblock URL \url{https://arxiv.org/abs/2202.05262}.

\bibitem[Pearl(2022)]{10.1145/3501714.3501736}
Judea Pearl.
\newblock \emph{Direct and Indirect Effects}, page 373–392.
\newblock Association for Computing Machinery, New York, NY, USA, 1 edition, 2022.
\newblock ISBN 9781450395861.
\newblock URL \url{https://doi.org/10.1145/3501714.3501736}.

\bibitem[Vig et~al.(2020)Vig, Gehrmann, Belinkov, Qian, Nevo, Singer, and Shieber]{NEURIPS2020_92650b2e}
Jesse Vig, Sebastian Gehrmann, Yonatan Belinkov, Sharon Qian, Daniel Nevo, Yaron Singer, and Stuart Shieber.
\newblock Investigating gender bias in language models using causal mediation analysis.
\newblock In H.~Larochelle, M.~Ranzato, R.~Hadsell, M.F. Balcan, and H.~Lin, editors, \emph{Advances in Neural Information Processing Systems}, volume~33, pages 12388--12401. Curran Associates, Inc., 2020.
\newblock URL \url{https://proceedings.neurips.cc/paper_files/paper/2020/file/92650b2e92217715fe312e6fa7b90d82-Paper.pdf}.

\bibitem[Kramár et~al.(2024)Kramár, Lieberum, Shah, and Nanda]{kramár2024atpefficientscalablemethod}
János Kramár, Tom Lieberum, Rohin Shah, and Neel Nanda.
\newblock Atp*: An efficient and scalable method for localizing llm behaviour to components, 2024.
\newblock URL \url{https://arxiv.org/abs/2403.00745}.

\bibitem[Bricken et~al.(2023)Bricken, Templeton, Batson, Chen, Jermyn, Conerly, Turner, Anil, Denison, Askell, Lasenby, Wu, Kravec, Schiefer, Maxwell, Joseph, Hatfield-Dodds, Tamkin, Nguyen, McLean, Burke, Hume, Carter, Henighan, and Olah]{bricken2023monosemanticity}
Trenton Bricken, Adly Templeton, Joshua Batson, Brian Chen, Adam Jermyn, Tom Conerly, Nick Turner, Cem Anil, Carson Denison, Amanda Askell, Robert Lasenby, Yifan Wu, Shauna Kravec, Nicholas Schiefer, Tim Maxwell, Nicholas Joseph, Zac Hatfield-Dodds, Alex Tamkin, Karina Nguyen, Brayden McLean, Josiah~E Burke, Tristan Hume, Shan Carter, Tom Henighan, and Christopher Olah.
\newblock Towards monosemanticity: Decomposing language models with dictionary learning.
\newblock \emph{Transformer Circuits Thread}, 2023.
\newblock https://transformer-circuits.pub/2023/monosemantic-features/index.html.

\bibitem[Marks et~al.(2025)Marks, Rager, Michaud, Belinkov, Bau, and Mueller]{marks2025sparsefeaturecircuitsdiscovering}
Samuel Marks, Can Rager, Eric~J. Michaud, Yonatan Belinkov, David Bau, and Aaron Mueller.
\newblock Sparse feature circuits: Discovering and editing interpretable causal graphs in language models, 2025.
\newblock URL \url{https://arxiv.org/abs/2403.19647}.

\bibitem[Ghorbani and Zou(2020)]{ghorbani2020neuronshapleydiscoveringresponsible}
Amirata Ghorbani and James Zou.
\newblock Neuron shapley: Discovering the responsible neurons, 2020.
\newblock URL \url{https://arxiv.org/abs/2002.09815}.

\bibitem[Cao et~al.(2021)Cao, Sanh, and Rush]{cao2021lowcomplexityprobingfindingsubnetworks}
Steven Cao, Victor Sanh, and Alexander~M. Rush.
\newblock Low-complexity probing via finding subnetworks, 2021.
\newblock URL \url{https://arxiv.org/abs/2104.03514}.

\bibitem[Csordás et~al.(2021)Csordás, van Steenkiste, and Schmidhuber]{csordás2021neuralnetsmodularinspecting}
Róbert Csordás, Sjoerd van Steenkiste, and Jürgen Schmidhuber.
\newblock Are neural nets modular? inspecting functional modularity through differentiable weight masks, 2021.
\newblock URL \url{https://arxiv.org/abs/2010.02066}.

\bibitem[Srivastava et~al.(2023)Srivastava, Rastogi, Rao, Shoeb, Abid, Fisch, Brown, Santoro, Gupta, Garriga-Alonso, Kluska, Lewkowycz, Agarwal, Power, Ray, Warstadt, Kocurek, Safaya, Tazarv, Xiang, Parrish, Nie, Hussain, Askell, Dsouza, Slone, Rahane, Iyer, Andreassen, Madotto, Santilli, Stuhlmüller, Dai, La, Lampinen, Zou, Jiang, Chen, Vuong, Gupta, Gottardi, Norelli, Venkatesh, Gholamidavoodi, Tabassum, Menezes, Kirubarajan, Mullokandov, Sabharwal, Herrick, Efrat, Erdem, Karakaş, Roberts, Loe, Zoph, Bojanowski, Özyurt, Hedayatnia, Neyshabur, Inden, Stein, Ekmekci, Lin, Howald, Orinion, Diao, Dour, Stinson, Argueta, Ramírez, Singh, Rathkopf, Meng, Baral, Wu, Callison-Burch, Waites, Voigt, Manning, Potts, Ramirez, Rivera, Siro, Raffel, Ashcraft, Garbacea, Sileo, Garrette, Hendrycks, Kilman, Roth, Freeman, Khashabi, Levy, González, Perszyk, Hernandez, Chen, Ippolito, Gilboa, Dohan, Drakard, Jurgens, Datta, Ganguli, Emelin, Kleyko, Yuret, Chen, Tam, Hupkes, Misra, Buzan, Mollo, Yang, Lee, Schrader,
  Shutova, Cubuk, Segal, Hagerman, Barnes, Donoway, Pavlick, Rodola, Lam, Chu, Tang, Erdem, Chang, Chi, Dyer, Jerzak, Kim, Manyasi, Zheltonozhskii, Xia, Siar, Martínez-Plumed, Happé, Chollet, Rong, Mishra, Winata, de~Melo, Kruszewski, Parascandolo, Mariani, Wang, Jaimovitch-López, Betz, Gur-Ari, Galijasevic, Kim, Rashkin, Hajishirzi, Mehta, Bogar, Shevlin, Schütze, Yakura, Zhang, Wong, Ng, Noble, Jumelet, Geissinger, Kernion, Hilton, Lee, Fisac, Simon, Koppel, Zheng, Zou, Kocoń, Thompson, Wingfield, Kaplan, Radom, Sohl-Dickstein, Phang, Wei, Yosinski, Novikova, Bosscher, Marsh, Kim, Taal, Engel, Alabi, Xu, Song, Tang, Waweru, Burden, Miller, Balis, Batchelder, Berant, Frohberg, Rozen, Hernandez-Orallo, Boudeman, Guerr, Jones, Tenenbaum, Rule, Chua, Kanclerz, Livescu, Krauth, Gopalakrishnan, Ignatyeva, Markert, Dhole, Gimpel, Omondi, Mathewson, Chiafullo, Shkaruta, Shridhar, McDonell, Richardson, Reynolds, Gao, Zhang, Dugan, Qin, Contreras-Ochando, Morency, Moschella, Lam, Noble, Schmidt, He, Colón,
  Metz, Şenel, Bosma, Sap, ter Hoeve, Farooqi, Faruqui, Mazeika, Baturan, Marelli, Maru, Quintana, Tolkiehn, Giulianelli, Lewis, Potthast, Leavitt, Hagen, Schubert, Baitemirova, Arnaud, McElrath, Yee, Cohen, Gu, Ivanitskiy, Starritt, Strube, Swędrowski, Bevilacqua, Yasunaga, Kale, Cain, Xu, Suzgun, Walker, Tiwari, Bansal, Aminnaseri, Geva, Gheini, T, Peng, Chi, Lee, Krakover, Cameron, Roberts, Doiron, Martinez, Nangia, Deckers, Muennighoff, Keskar, Iyer, Constant, Fiedel, Wen, Zhang, Agha, Elbaghdadi, Levy, Evans, Casares, Doshi, Fung, Liang, Vicol, Alipoormolabashi, Liao, Liang, Chang, Eckersley, Htut, Hwang, Miłkowski, Patil, Pezeshkpour, Oli, Mei, Lyu, Chen, Banjade, Rudolph, Gabriel, Habacker, Risco, Millière, Garg, Barnes, Saurous, Arakawa, Raymaekers, Frank, Sikand, Novak, Sitelew, LeBras, Liu, Jacobs, Zhang, Salakhutdinov, Chi, Lee, Stovall, Teehan, Yang, Singh, Mohammad, Anand, Dillavou, Shleifer, Wiseman, Gruetter, Bowman, Schoenholz, Han, Kwatra, Rous, Ghazarian, Ghosh, Casey, Bischoff,
  Gehrmann, Schuster, Sadeghi, Hamdan, Zhou, Srivastava, Shi, Singh, Asaadi, Gu, Pachchigar, Toshniwal, Upadhyay, Shyamolima, Debnath, Shakeri, Thormeyer, Melzi, Reddy, Makini, Lee, Torene, Hatwar, Dehaene, Divic, Ermon, Biderman, Lin, Prasad, Piantadosi, Shieber, Misherghi, Kiritchenko, Mishra, Linzen, Schuster, Li, Yu, Ali, Hashimoto, Wu, Desbordes, Rothschild, Phan, Wang, Nkinyili, Schick, Kornev, Tunduny, Gerstenberg, Chang, Neeraj, Khot, Shultz, Shaham, Misra, Demberg, Nyamai, Raunak, Ramasesh, Prabhu, Padmakumar, Srikumar, Fedus, Saunders, Zhang, Vossen, Ren, Tong, Zhao, Wu, Shen, Yaghoobzadeh, Lakretz, Song, Bahri, Choi, Yang, Hao, Chen, Belinkov, Hou, Hou, Bai, Seid, Zhao, Wang, Wang, Wang, and Wu]{srivastava2023imitationgamequantifyingextrapolating}
Aarohi Srivastava, Abhinav Rastogi, Abhishek Rao, Abu Awal~Md Shoeb, Abubakar Abid, Adam Fisch, Adam~R. Brown, Adam Santoro, Aditya Gupta, Adrià Garriga-Alonso, Agnieszka Kluska, Aitor Lewkowycz, Akshat Agarwal, Alethea Power, Alex Ray, Alex Warstadt, Alexander~W. Kocurek, Ali Safaya, Ali Tazarv, Alice Xiang, Alicia Parrish, Allen Nie, Aman Hussain, Amanda Askell, Amanda Dsouza, Ambrose Slone, Ameet Rahane, Anantharaman~S. Iyer, Anders Andreassen, Andrea Madotto, Andrea Santilli, Andreas Stuhlmüller, Andrew Dai, Andrew La, Andrew Lampinen, Andy Zou, Angela Jiang, Angelica Chen, Anh Vuong, Animesh Gupta, Anna Gottardi, Antonio Norelli, Anu Venkatesh, Arash Gholamidavoodi, Arfa Tabassum, Arul Menezes, Arun Kirubarajan, Asher Mullokandov, Ashish Sabharwal, Austin Herrick, Avia Efrat, Aykut Erdem, Ayla Karakaş, B.~Ryan Roberts, Bao~Sheng Loe, Barret Zoph, Bartłomiej Bojanowski, Batuhan Özyurt, Behnam Hedayatnia, Behnam Neyshabur, Benjamin Inden, Benno Stein, Berk Ekmekci, Bill~Yuchen Lin, Blake Howald, Bryan
  Orinion, Cameron Diao, Cameron Dour, Catherine Stinson, Cedrick Argueta, César~Ferri Ramírez, Chandan Singh, Charles Rathkopf, Chenlin Meng, Chitta Baral, Chiyu Wu, Chris Callison-Burch, Chris Waites, Christian Voigt, Christopher~D. Manning, Christopher Potts, Cindy Ramirez, Clara~E. Rivera, Clemencia Siro, Colin Raffel, Courtney Ashcraft, Cristina Garbacea, Damien Sileo, Dan Garrette, Dan Hendrycks, Dan Kilman, Dan Roth, Daniel Freeman, Daniel Khashabi, Daniel Levy, Daniel~Moseguí González, Danielle Perszyk, Danny Hernandez, Danqi Chen, Daphne Ippolito, Dar Gilboa, David Dohan, David Drakard, David Jurgens, Debajyoti Datta, Deep Ganguli, Denis Emelin, Denis Kleyko, Deniz Yuret, Derek Chen, Derek Tam, Dieuwke Hupkes, Diganta Misra, Dilyar Buzan, Dimitri~Coelho Mollo, Diyi Yang, Dong-Ho Lee, Dylan Schrader, Ekaterina Shutova, Ekin~Dogus Cubuk, Elad Segal, Eleanor Hagerman, Elizabeth Barnes, Elizabeth Donoway, Ellie Pavlick, Emanuele Rodola, Emma Lam, Eric Chu, Eric Tang, Erkut Erdem, Ernie Chang,
  Ethan~A. Chi, Ethan Dyer, Ethan Jerzak, Ethan Kim, Eunice~Engefu Manyasi, Evgenii Zheltonozhskii, Fanyue Xia, Fatemeh Siar, Fernando Martínez-Plumed, Francesca Happé, Francois Chollet, Frieda Rong, Gaurav Mishra, Genta~Indra Winata, Gerard de~Melo, Germán Kruszewski, Giambattista Parascandolo, Giorgio Mariani, Gloria Wang, Gonzalo Jaimovitch-López, Gregor Betz, Guy Gur-Ari, Hana Galijasevic, Hannah Kim, Hannah Rashkin, Hannaneh Hajishirzi, Harsh Mehta, Hayden Bogar, Henry Shevlin, Hinrich Schütze, Hiromu Yakura, Hongming Zhang, Hugh~Mee Wong, Ian Ng, Isaac Noble, Jaap Jumelet, Jack Geissinger, Jackson Kernion, Jacob Hilton, Jaehoon Lee, Jaime~Fernández Fisac, James~B. Simon, James Koppel, James Zheng, James Zou, Jan Kocoń, Jana Thompson, Janelle Wingfield, Jared Kaplan, Jarema Radom, Jascha Sohl-Dickstein, Jason Phang, Jason Wei, Jason Yosinski, Jekaterina Novikova, Jelle Bosscher, Jennifer Marsh, Jeremy Kim, Jeroen Taal, Jesse Engel, Jesujoba Alabi, Jiacheng Xu, Jiaming Song, Jillian Tang, Joan
  Waweru, John Burden, John Miller, John~U. Balis, Jonathan Batchelder, Jonathan Berant, Jörg Frohberg, Jos Rozen, Jose Hernandez-Orallo, Joseph Boudeman, Joseph Guerr, Joseph Jones, Joshua~B. Tenenbaum, Joshua~S. Rule, Joyce Chua, Kamil Kanclerz, Karen Livescu, Karl Krauth, Karthik Gopalakrishnan, Katerina Ignatyeva, Katja Markert, Kaustubh~D. Dhole, Kevin Gimpel, Kevin Omondi, Kory Mathewson, Kristen Chiafullo, Ksenia Shkaruta, Kumar Shridhar, Kyle McDonell, Kyle Richardson, Laria Reynolds, Leo Gao, Li~Zhang, Liam Dugan, Lianhui Qin, Lidia Contreras-Ochando, Louis-Philippe Morency, Luca Moschella, Lucas Lam, Lucy Noble, Ludwig Schmidt, Luheng He, Luis~Oliveros Colón, Luke Metz, Lütfi~Kerem Şenel, Maarten Bosma, Maarten Sap, Maartje ter Hoeve, Maheen Farooqi, Manaal Faruqui, Mantas Mazeika, Marco Baturan, Marco Marelli, Marco Maru, Maria Jose~Ramírez Quintana, Marie Tolkiehn, Mario Giulianelli, Martha Lewis, Martin Potthast, Matthew~L. Leavitt, Matthias Hagen, Mátyás Schubert, Medina~Orduna
  Baitemirova, Melody Arnaud, Melvin McElrath, Michael~A. Yee, Michael Cohen, Michael Gu, Michael Ivanitskiy, Michael Starritt, Michael Strube, Michał Swędrowski, Michele Bevilacqua, Michihiro Yasunaga, Mihir Kale, Mike Cain, Mimee Xu, Mirac Suzgun, Mitch Walker, Mo~Tiwari, Mohit Bansal, Moin Aminnaseri, Mor Geva, Mozhdeh Gheini, Mukund~Varma T, Nanyun Peng, Nathan~A. Chi, Nayeon Lee, Neta Gur-Ari Krakover, Nicholas Cameron, Nicholas Roberts, Nick Doiron, Nicole Martinez, Nikita Nangia, Niklas Deckers, Niklas Muennighoff, Nitish~Shirish Keskar, Niveditha~S. Iyer, Noah Constant, Noah Fiedel, Nuan Wen, Oliver Zhang, Omar Agha, Omar Elbaghdadi, Omer Levy, Owain Evans, Pablo Antonio~Moreno Casares, Parth Doshi, Pascale Fung, Paul~Pu Liang, Paul Vicol, Pegah Alipoormolabashi, Peiyuan Liao, Percy Liang, Peter Chang, Peter Eckersley, Phu~Mon Htut, Pinyu Hwang, Piotr Miłkowski, Piyush Patil, Pouya Pezeshkpour, Priti Oli, Qiaozhu Mei, Qing Lyu, Qinlang Chen, Rabin Banjade, Rachel~Etta Rudolph, Raefer Gabriel, Rahel
  Habacker, Ramon Risco, Raphaël Millière, Rhythm Garg, Richard Barnes, Rif~A. Saurous, Riku Arakawa, Robbe Raymaekers, Robert Frank, Rohan Sikand, Roman Novak, Roman Sitelew, Ronan LeBras, Rosanne Liu, Rowan Jacobs, Rui Zhang, Ruslan Salakhutdinov, Ryan Chi, Ryan Lee, Ryan Stovall, Ryan Teehan, Rylan Yang, Sahib Singh, Saif~M. Mohammad, Sajant Anand, Sam Dillavou, Sam Shleifer, Sam Wiseman, Samuel Gruetter, Samuel~R. Bowman, Samuel~S. Schoenholz, Sanghyun Han, Sanjeev Kwatra, Sarah~A. Rous, Sarik Ghazarian, Sayan Ghosh, Sean Casey, Sebastian Bischoff, Sebastian Gehrmann, Sebastian Schuster, Sepideh Sadeghi, Shadi Hamdan, Sharon Zhou, Shashank Srivastava, Sherry Shi, Shikhar Singh, Shima Asaadi, Shixiang~Shane Gu, Shubh Pachchigar, Shubham Toshniwal, Shyam Upadhyay, Shyamolima, Debnath, Siamak Shakeri, Simon Thormeyer, Simone Melzi, Siva Reddy, Sneha~Priscilla Makini, Soo-Hwan Lee, Spencer Torene, Sriharsha Hatwar, Stanislas Dehaene, Stefan Divic, Stefano Ermon, Stella Biderman, Stephanie Lin, Stephen
  Prasad, Steven~T. Piantadosi, Stuart~M. Shieber, Summer Misherghi, Svetlana Kiritchenko, Swaroop Mishra, Tal Linzen, Tal Schuster, Tao Li, Tao Yu, Tariq Ali, Tatsu Hashimoto, Te-Lin Wu, Théo Desbordes, Theodore Rothschild, Thomas Phan, Tianle Wang, Tiberius Nkinyili, Timo Schick, Timofei Kornev, Titus Tunduny, Tobias Gerstenberg, Trenton Chang, Trishala Neeraj, Tushar Khot, Tyler Shultz, Uri Shaham, Vedant Misra, Vera Demberg, Victoria Nyamai, Vikas Raunak, Vinay Ramasesh, Vinay~Uday Prabhu, Vishakh Padmakumar, Vivek Srikumar, William Fedus, William Saunders, William Zhang, Wout Vossen, Xiang Ren, Xiaoyu Tong, Xinran Zhao, Xinyi Wu, Xudong Shen, Yadollah Yaghoobzadeh, Yair Lakretz, Yangqiu Song, Yasaman Bahri, Yejin Choi, Yichi Yang, Yiding Hao, Yifu Chen, Yonatan Belinkov, Yu~Hou, Yufang Hou, Yuntao Bai, Zachary Seid, Zhuoye Zhao, Zijian Wang, Zijie~J. Wang, Zirui Wang, and Ziyi Wu.
\newblock Beyond the imitation game: Quantifying and extrapolating the capabilities of language models, 2023.
\newblock URL \url{https://arxiv.org/abs/2206.04615}.

\bibitem[Nainani et~al.(2024)Nainani, Vaidyanathan, Yeung, Gupta, and Jensen]{nainani2024adaptivecircuitbehaviorgeneralization}
Jatin Nainani, Sankaran Vaidyanathan, AJ~Yeung, Kartik Gupta, and David Jensen.
\newblock Adaptive circuit behavior and generalization in mechanistic interpretability, 2024.
\newblock URL \url{https://arxiv.org/abs/2411.16105}.

\bibitem[Conmy et~al.(2023)Conmy, Mavor-Parker, Lynch, Heimersheim, and Garriga-Alonso]{conmy2023automatedcircuitdiscoverymechanistic}
Arthur Conmy, Augustine~N. Mavor-Parker, Aengus Lynch, Stefan Heimersheim, and Adrià Garriga-Alonso.
\newblock Towards automated circuit discovery for mechanistic interpretability, 2023.
\newblock URL \url{https://arxiv.org/abs/2304.14997}.

\bibitem[Eisape et~al.(2024)Eisape, Tessler, Dasgupta, Sha, van Steenkiste, and Linzen]{eisape2024systematiccomparisonsyllogisticreasoning}
Tiwalayo Eisape, MH~Tessler, Ishita Dasgupta, Fei Sha, Sjoerd van Steenkiste, and Tal Linzen.
\newblock A systematic comparison of syllogistic reasoning in humans and language models, 2024.
\newblock URL \url{https://arxiv.org/abs/2311.00445}.

\bibitem[Ando et~al.(2023)Ando, Morishita, Abe, Mineshima, and Okada]{ando2023evaluatinglargelanguagemodels}
Risako Ando, Takanobu Morishita, Hirohiko Abe, Koji Mineshima, and Mitsuhiro Okada.
\newblock Evaluating large language models with neubaroco: Syllogistic reasoning ability and human-like biases, 2023.
\newblock URL \url{https://arxiv.org/abs/2306.12567}.

\bibitem[Kim et~al.(2025)Kim, Valentino, and Freitas]{kim2025mechanisticinterpretationsyllogisticreasoning}
Geonhee Kim, Marco Valentino, and André Freitas.
\newblock A mechanistic interpretation of syllogistic reasoning in auto-regressive language models, 2025.
\newblock URL \url{https://arxiv.org/abs/2408.08590}.

\end{thebibliography}

\clearpage
\appendix
\section{Syllogism Dataset Construction} \label{syllogism_dataset_construction}
Syllogistic prompts were created using templates. \texttt{[TRUTH\_VALUE]}~$\in \{\texttt{true}, \texttt{false}\}$, and \texttt{[A], [B], [C]} are sampled from capital letters. See table \ref{table:construction} for templates of each syllogism format

\begin{table*}[h]
\centering
\renewcommand{\arraystretch}{1.4}
\begin{tabularx}{\textwidth}{>{\raggedright\arraybackslash}p{2.6cm}X}
\toprule
\textbf{Type} & \textbf{Template} \\
\midrule

\textbf{Simple Syllogism} & 
1. Statement [A] is [TRUTH\_VALUE\_1]. Statement [B] has the same truth value as [A]. Statement [B] is [TRUTH\_VALUE\_1]. \\
& 2. Statement [A] is [TRUTH\_VALUE\_1]. Statement [B] matches statement A. Statement B is [TRUTH\_VALUE\_1]. \\
& 3. \textit{(Extended)} Statement [A] is [TRUTH\_VALUE\_1]. Statement [B] must match [A]. Statement [C] doesn’t matter. Statement [B] is [TRUTH\_VALUE\_1]. \\

\cmidrule(l){1-2}

\textbf{Opposite Syllogism} & 
1. Statement [B] has the opposite truth value of [A]. Statement [A] is [TRUTH\_VALUE\_1]. Statement [B] is [TRUTH\_VALUE\_2]. \\
& 2. Statement [A] and statement [B] are opposites. Statement [A] is [TRUTH\_VALUE\_1]. Statement [B] is [TRUTH\_VALUE\_2]. \\

\cmidrule(l){1-2}

\textbf{Complex Syllogism} & 
1. Statement [A] is [TRUTH\_VALUE\_1]. Statement [B] has same truth value as [A]. Statement [C] is [TRUTH\_VALUE\_2]. Statement [B] is [TRUTH\_VALUE\_3]. \\
&\textit{(Harder constraint)}: [TRUTH\_VALUE\_2] = $\neg$[TRUTH\_VALUE\_1]. \\

\cmidrule(l){1-2}

\textbf{Complex Opposite Syllogism} & 
1. Statement [A] is [TRUTH\_VALUE\_1]. Statement [B] has the opposite truth value of [A]. Statement [C] is [TRUTH\_VALUE\_2]. Statement [B] is [TRUTH\_VALUE\_3]. \\
& 2. Statement [A] and [B] are opposites. Statement [C] has the same truth value as [A]. Statement [B] is [TRUTH\_VALUE\_3]. \\
& 3. Statement [A] is [TRUTH\_VALUE\_1]. Statement [A] and [B] are opposites. Statement [C] is [TRUTH\_VALUE\_2]. Statement [B] is [TRUTH\_VALUE\_3]. \\

\bottomrule
\end{tabularx}
\caption{Templates used for generating syllogistic prompts.}
\label{table:construction}
\end{table*}

\section{Generality Across Binary Contrasts} \label{generality_of_binary_task}

Having established mechanistic evidence for circuits supporting binary truth tasks in both the simple and opposite syllogism settings, we next evaluate the generality of these circuits beyond the original \texttt{true}/\texttt{false} framing. Specifically, we test whether the same circuits generalize to analogous binary pairs: \texttt{(right, wrong)}, \texttt{(good, bad)}, \texttt{(positive, negative)}, and \texttt{(correct, incorrect)}.

To do so, we apply both the simple syllogism circuit ($\bm{C}_{SS}$) and opposite syllogism circuit ($\bm{C}_{OS}$) to each pair and compare their performance to the full GPT-2 Small model. As shown by tables \ref{table:logit_diff_css} and \ref{table:logit_diff_cos} we find that the original circuits often match or even outperform the full model in logit difference between most binary pairs of tokens. This provides compelling evidence that the binary task is not specific to a particular token pair, but instead reflects a transferable reasoning mechanism.

\begin{figure*}[htbp]
    \centering
    \subfloat[Simple Syllogism with Right/Wrong]{\includegraphics[width=0.45\textwidth]{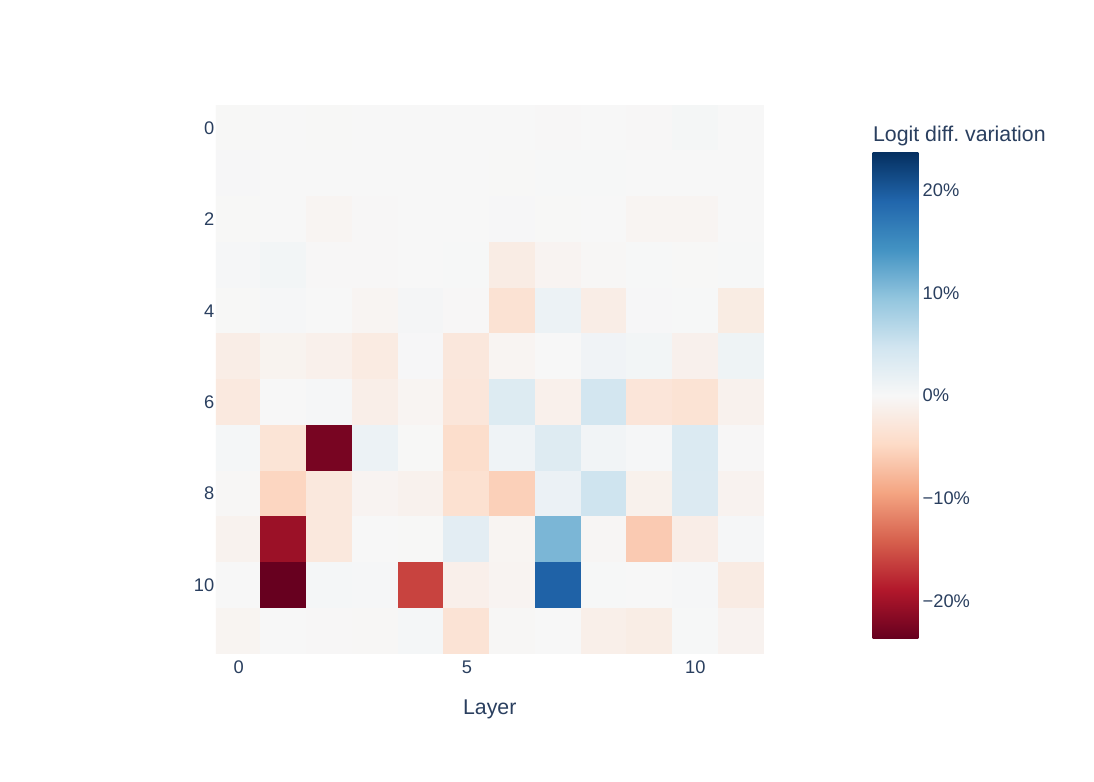}}
    \hfill
    \subfloat[Opposite Syllogism with Right/Wrong]{\includegraphics[width=0.45\textwidth]{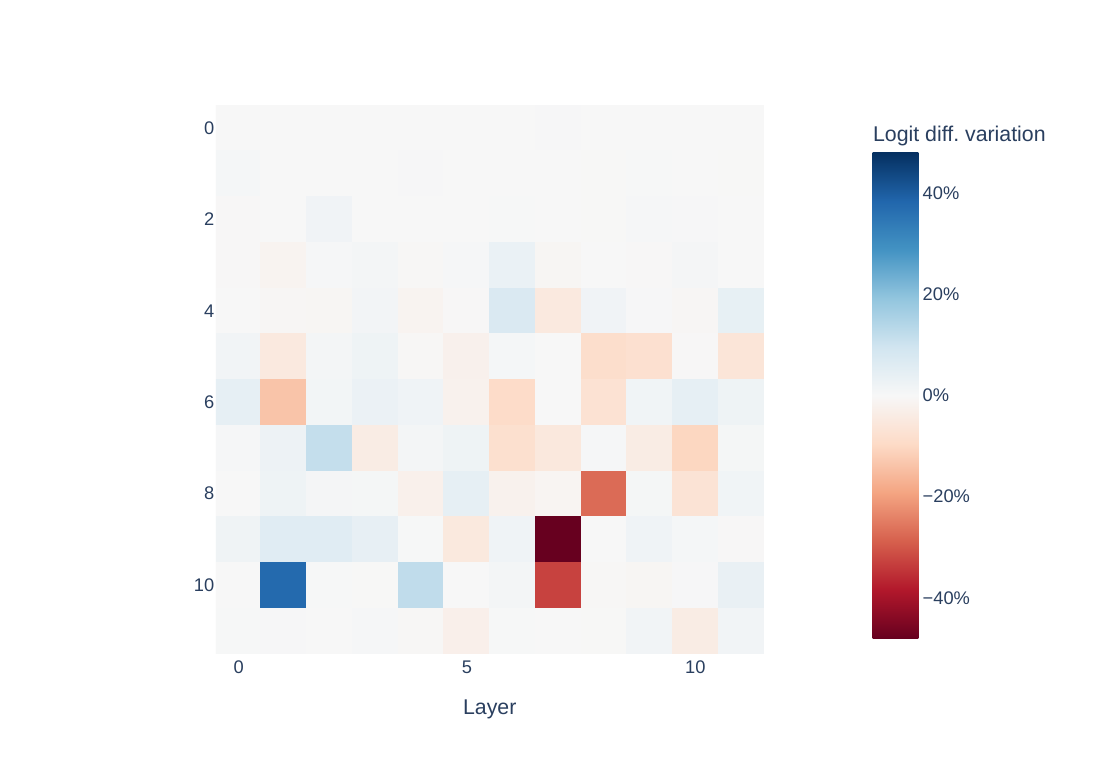}}
    \caption{Binary task results of Right/Wrong}
    \label{figure:binary_rw}
\end{figure*}

\begin{figure*}[htbp]
    \centering
    \subfloat[Simple Syllogism with Correct/Incorrect]{\includegraphics[width=0.45\textwidth]{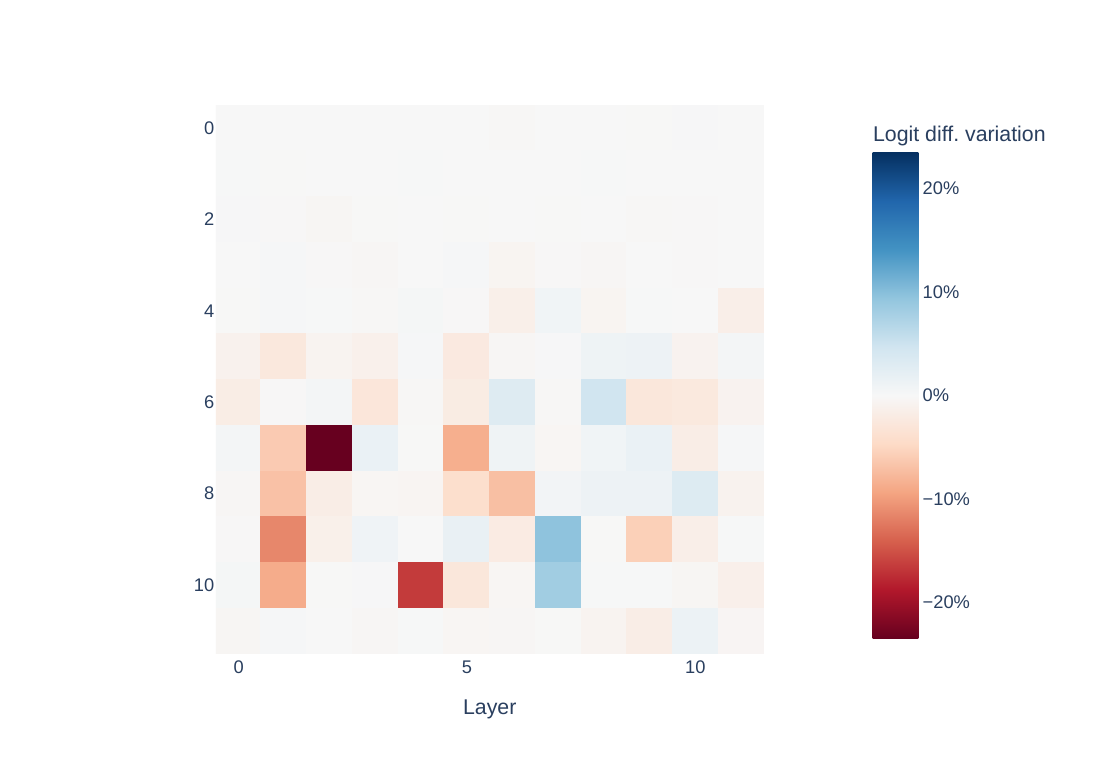}}
    \hfill
    \subfloat[Opposite Syllogism with Correct/Incorrect]{\includegraphics[width=0.45\textwidth]{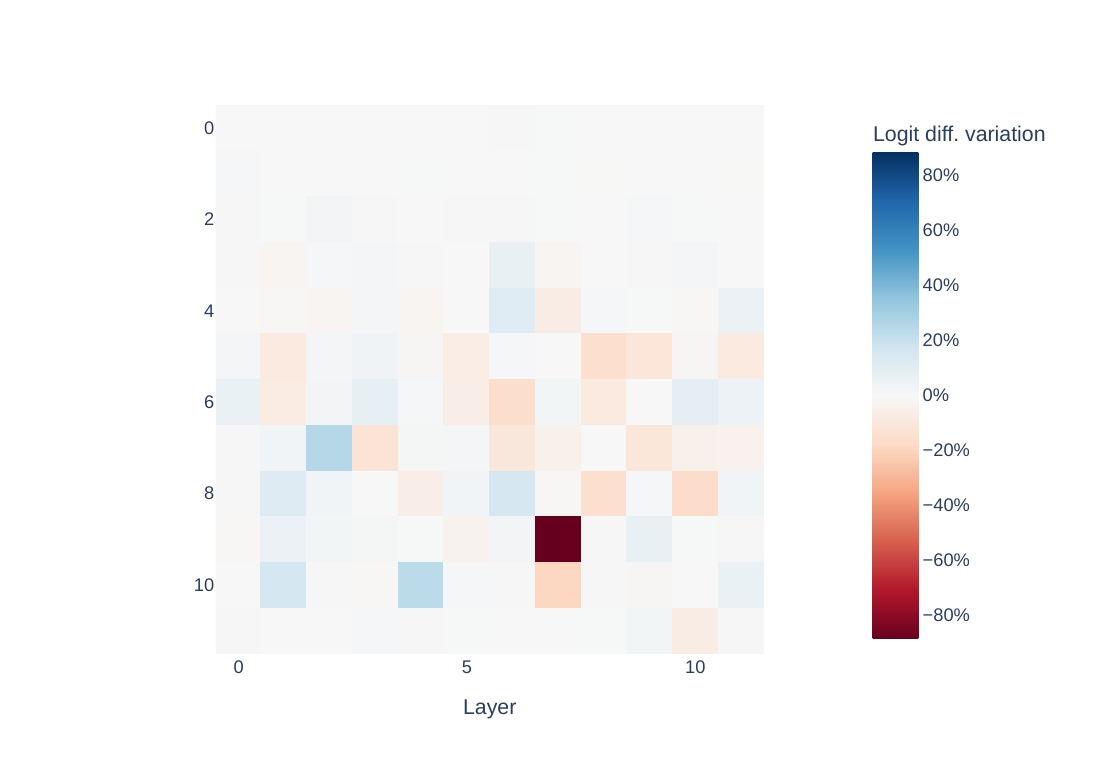}}
    \caption{Binary task results of Correct/Incorrect}
    \label{figure:binary_ci}
\end{figure*}

\begin{figure*}[htbp]
    \centering
    \subfloat[Simple Syllogism with Positive/Negative]{\includegraphics[width=0.45\textwidth]{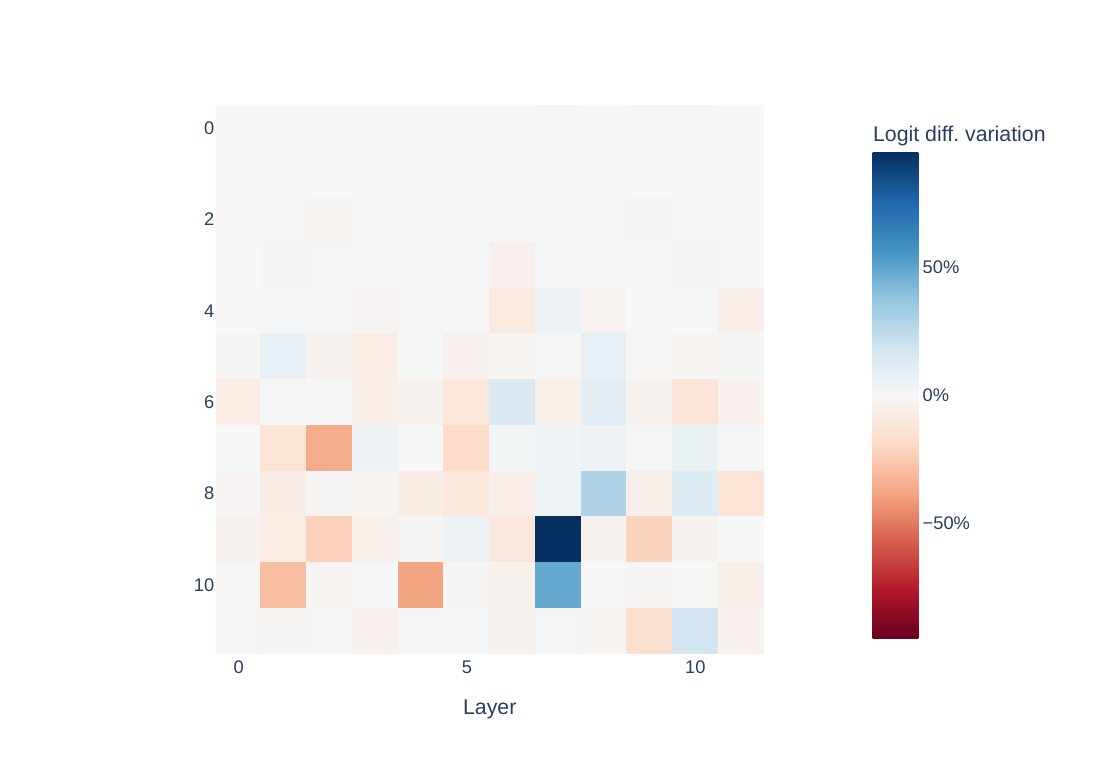}}
    \hfill
    \subfloat[Opposite Syllogism with Positive/Negative]{\includegraphics[width=0.45\textwidth]{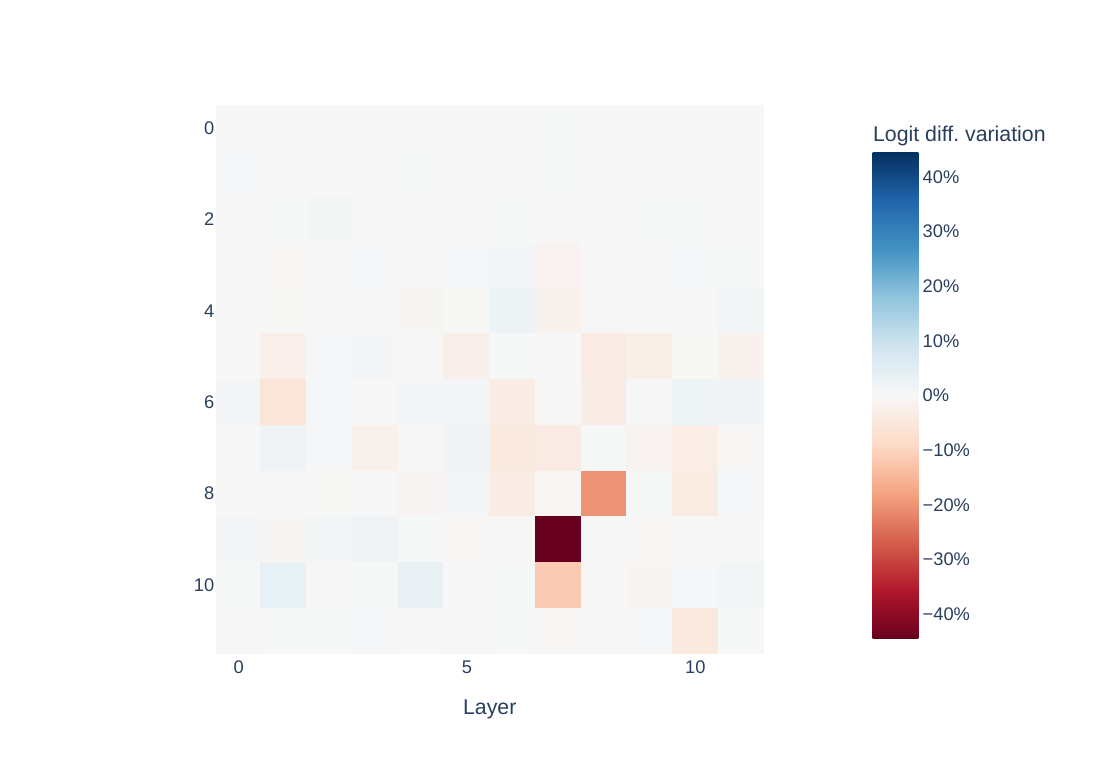}}
    \caption{Binary task results of Positive/Negative}
    \label{figure:binary_pn}
\end{figure*}

\begin{figure*}[htbp]
    \centering
    \subfloat[Simple Syllogism with Good/Bad]{\includegraphics[width=0.45\textwidth]{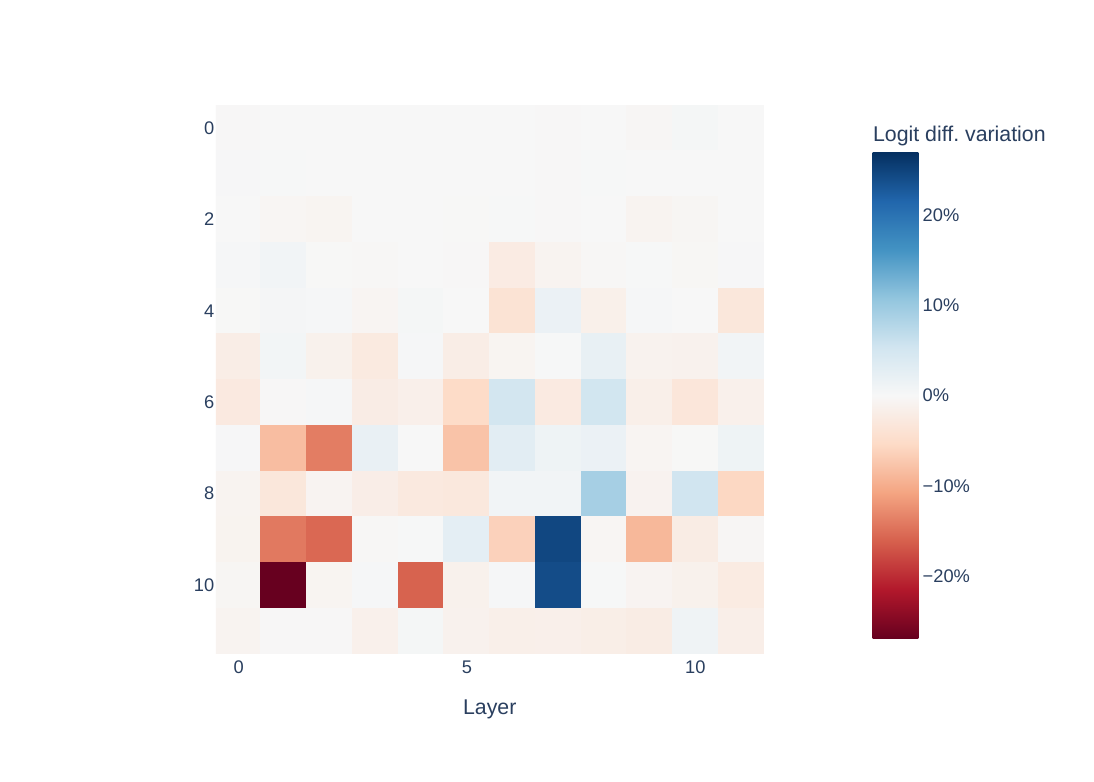}}
    \hfill
    \subfloat[Opposite Syllogism with Good/Bad]{\includegraphics[width=0.45\textwidth]{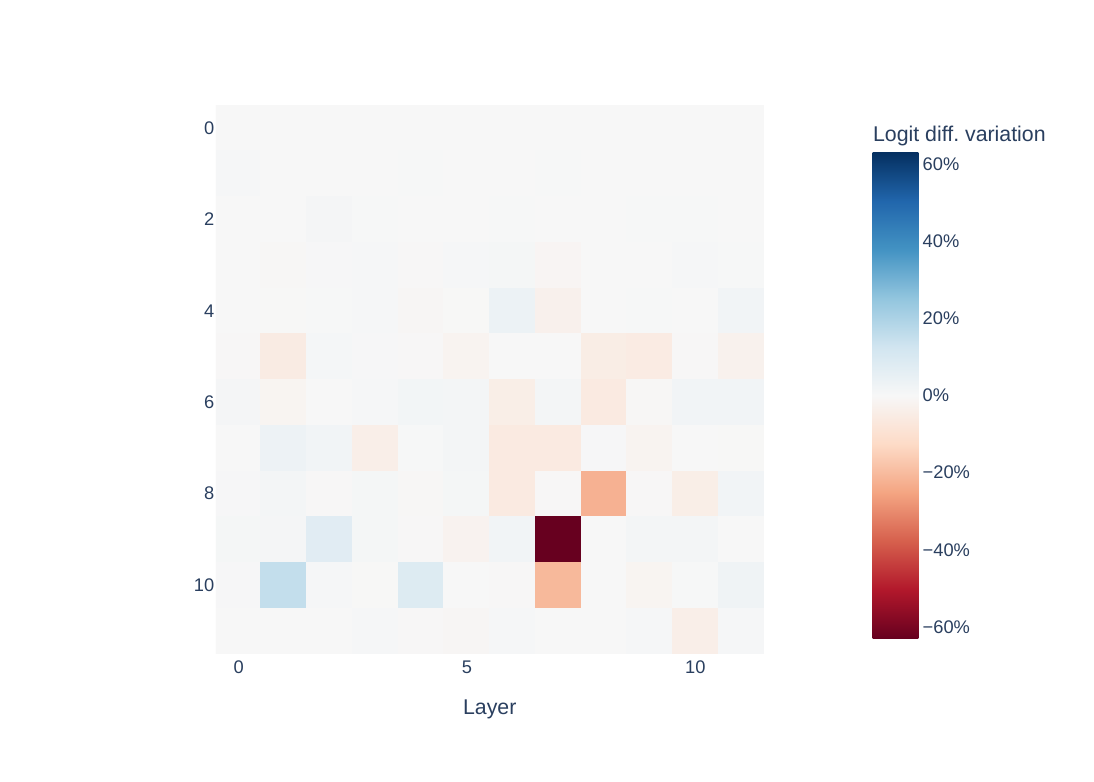}}
    \caption{Binary task results of Good/Bad}
    \label{figure:binary_gb}
\end{figure*}

To further validate generalization, we visualize direct path patching attention results across each binary pair. As seen in Figures~\ref{figure:binary_rw}–\ref{figure:binary_gb}, across the binary pairs of tokens, the core attention heads relevant to the simple and opposite syllogism cases are opposite in their effect on logit difference. 
\noindent
\begin{table*}[t]
\centering
\begin{tabularx}{\textwidth}{>{\centering\bfseries}Xccccc}
\toprule
 & \textbf{Original} & \textbf{Good/Bad} & \textbf{Pos/Neg} & \textbf{Correct/Incorrect} & \textbf{Right/Wrong} \\
\midrule
GPT-2 Small & 1.8399 & 1.7738 & 0.6958 & 2.1221 & 2.0309 \\
$\bm{C}_{SS}$ & 1.9234 & 1.9940 & 1.1584 & 1.6785 & 2.1599 \\
\bottomrule
\end{tabularx}
\caption{Transferability of $\bm{C}_{SS}$ to other binary token pairs }
\label{table:logit_diff_css}
\end{table*}
\noindent
\begin{table*}[t]
\centering
\begin{tabularx}{\textwidth}{>{\centering\bfseries}Xccccc}
\toprule
 & \textbf{Original} & \textbf{Good/Bad} & \textbf{Pos/Neg} & \textbf{Correct/Incorrect} & \textbf{Right/Wrong} \\
\midrule
GPT-2 Small & 1.2632 & 2.1163 & 3.0032 & 0.7986 & 1.3469 \\
$\bm{C}_{OS}$ & 1.3136 & 1.7136 & 1.0113 & 0.8142 & 1.2481 \\
\bottomrule
\end{tabularx}
\caption{Transferability of $\bm{C}_{OS}$ to other binary token pairs }
\label{table:logit_diff_cos}
\end{table*}

\section{Disentangling MLP Contributions via Patching}\label{do_mlps_matter_ss}

\begin{figure*}[h]
    \centering
    \subfloat[MLP effects with attention context\label{fig:mlp_w_attn_ss}]
    {\includegraphics[width=0.45\textwidth]{assets/simple/ss_mlp_direct_effect_w_attn_head.pdf}}
    \hfill
    \subfloat[MLP effects without attention context\label{fig:mlp_wo_attn_ss}]
    {\includegraphics[width=0.45\textwidth]{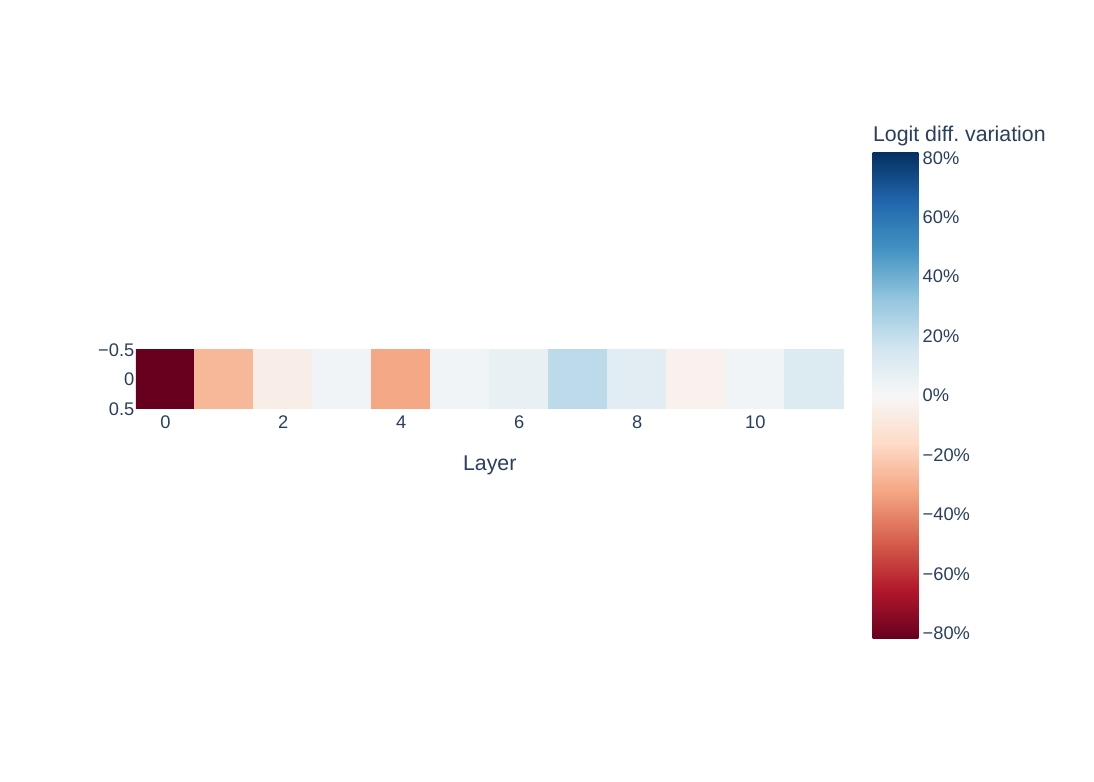}}
    \caption{Path patching MLPs in the simple syllogism task. (a) shows effects when MLPs are patched with attention context preserved. No MLPs have significant importance; (b) shows isolated MLP contributions without attention context. Early MLPs, specifically MLP0, appear relevant for the task}
    \label{figure:mlp_patch_comparison_ss}
\end{figure*}

To assess the contribution of MLPs to the model's output, we perform path patching both with and without attention restored. Figure \ref{fig:mlp_wo_attn_ss} shows that early-layer MLPs—particularly MLP0—appear to significantly affect the logits when patched in isolation. This aligns with prior observations that MLP0 functions as an extended embedding layer, especially when attention is absent \citep{mcdougall2023copysuppressioncomprehensivelyunderstanding, wang2022interpretabilitywildcircuitindirect}.

However, once attention is also restored, the influence of these early MLPs sharply diminishes. This suggests their apparent impact in the no-attention condition is largely an artifact of missing context, rather than a reflection of GPT2 semantic ability to complete syllogisms.

For this reason, in all subsequent experiments analyzing MLP effects, we report results with attention paths patched in. This allows us to isolate the true downstream influence of MLPs under more realistic model conditions.

\begin{figure*}[h]
    \centering
    \subfloat[MLP effects with attention context\label{fig:mlp_w_attn}]
    {\includegraphics[width=0.45\textwidth]{assets/opposite/os_mlp_direct_effect_w_attn_head.pdf}}
    \hfill
    \subfloat[MLP effects without attention context\label{fig:mlp_wo_attn}]
    {\includegraphics[width=0.45\textwidth]{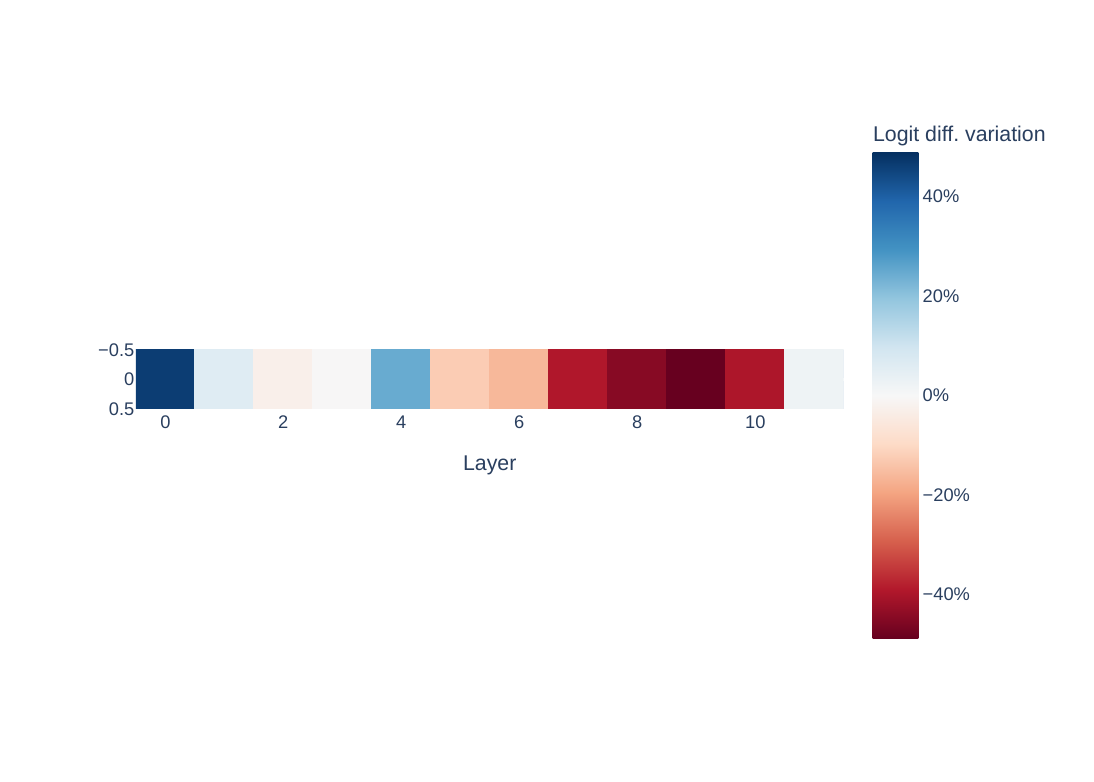}}
    \caption{Path patching MLPs in the opposite syllogism task. (a) shows effects when MLPs are patched with attention context preserved; (b) shows isolated MLP contributions without attention context.}
\end{figure*}

\section{Extension to Larger Models}
\label{section:larger_models}
To assess whether the findings observed in GPT-2 Small generalize across model scale and architecture, we extend our experiments to several larger models: GPT-2 XL, Pythia 1.4B, Qwen3-1.7B, and LLaMA3.2-1B.

Across all models, we continue to observe empirical signatures of binary behavior: heads relevant to the simple and opposite syllogism tasks tend to exert opposing effects on the logits. MLP layers remain important in the opposite syllogism task for all models except Pythia 1.4B, mirroring the behavior observed in GPT-2 Small. Notably, Table \ref{ald_across_models} shows that performance on the simple syllogism format degrades significantly in larger models, suggesting that task generalization does not uniformly scale with model size.

All models retain some attention heads exhibiting negative-copy behavior. However, the influence of these heads on output logits is more muted compared to GPT-2 Small. In particular, the heads most responsible for enabling opposite syllogism performance in the larger models are not the negative heads. Qwen3-1.7B, for instance, contains relatively few negative heads, and those it has do not drive logit differences in either task. An exception is Pythia 1.4B, whose success on the opposite task remains closely tied to the activity of its negative-copy heads.

Interestingly, across all models, the heads most influential on model output tend to exhibit strong induction behavior (e.g., ABA $\rightarrow$ B), regardless of whether they also contribute to the task-relevant distinction. Yet despite this variability in attention head dynamics, the consistent involvement of MLPs in the opposite task—and their near absence in the simple task—suggests a robust division of labor: negation appears to depend more heavily on the feedforward path than on attention alone. This may help constrain future hypotheses about the mechanistic implementation of logical inversion and contextual negation in transformer models.

\begin{figure*}[h]
    \centering
    \includegraphics[width=0.45\textwidth]{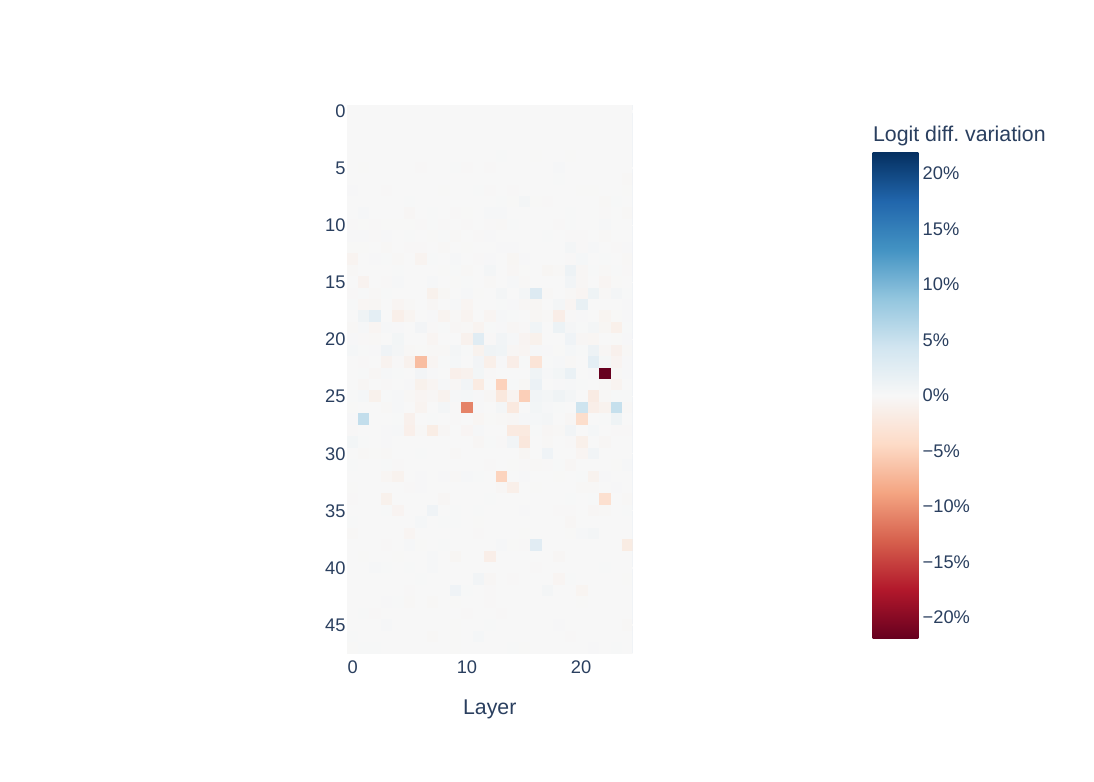}
    \includegraphics[width=0.45\textwidth]{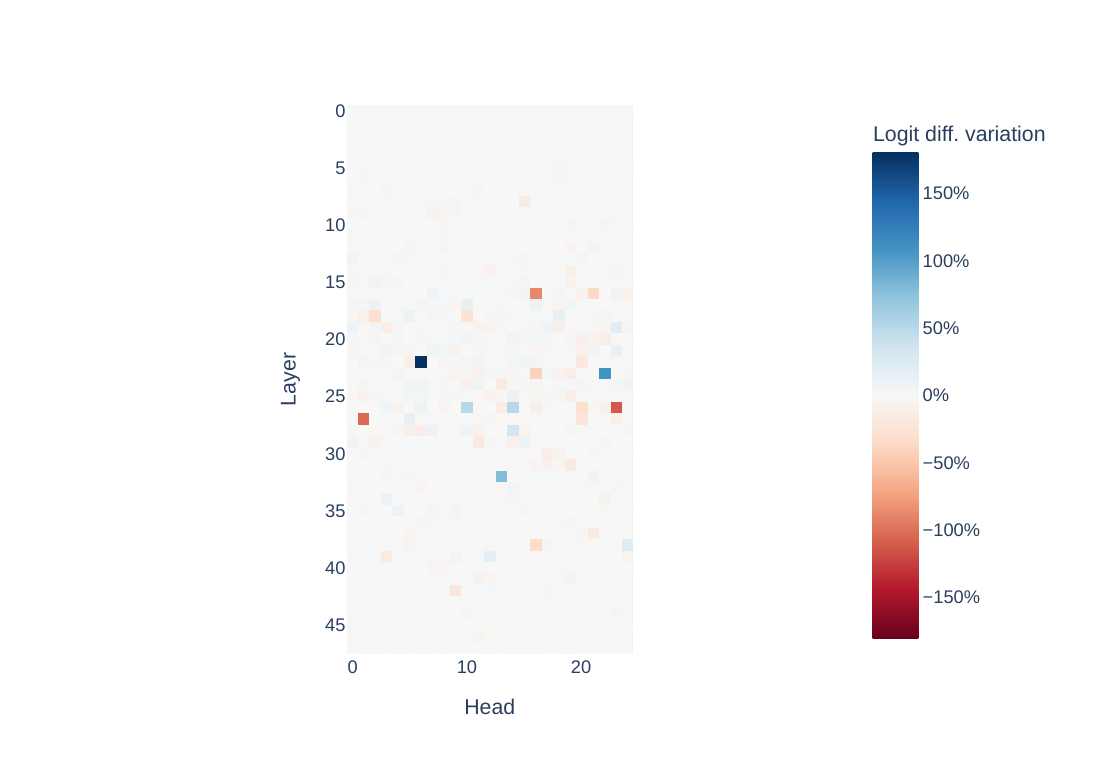}
    
    \includegraphics[width=0.45\textwidth]{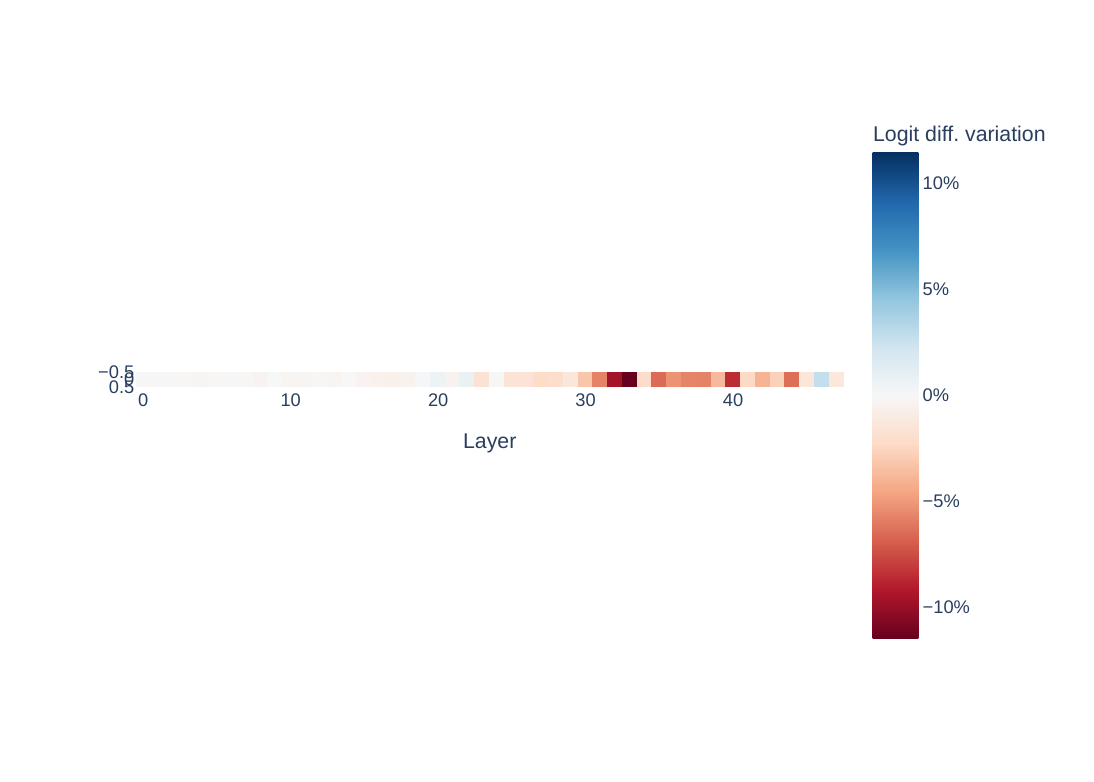}
    \includegraphics[width=0.45\textwidth]{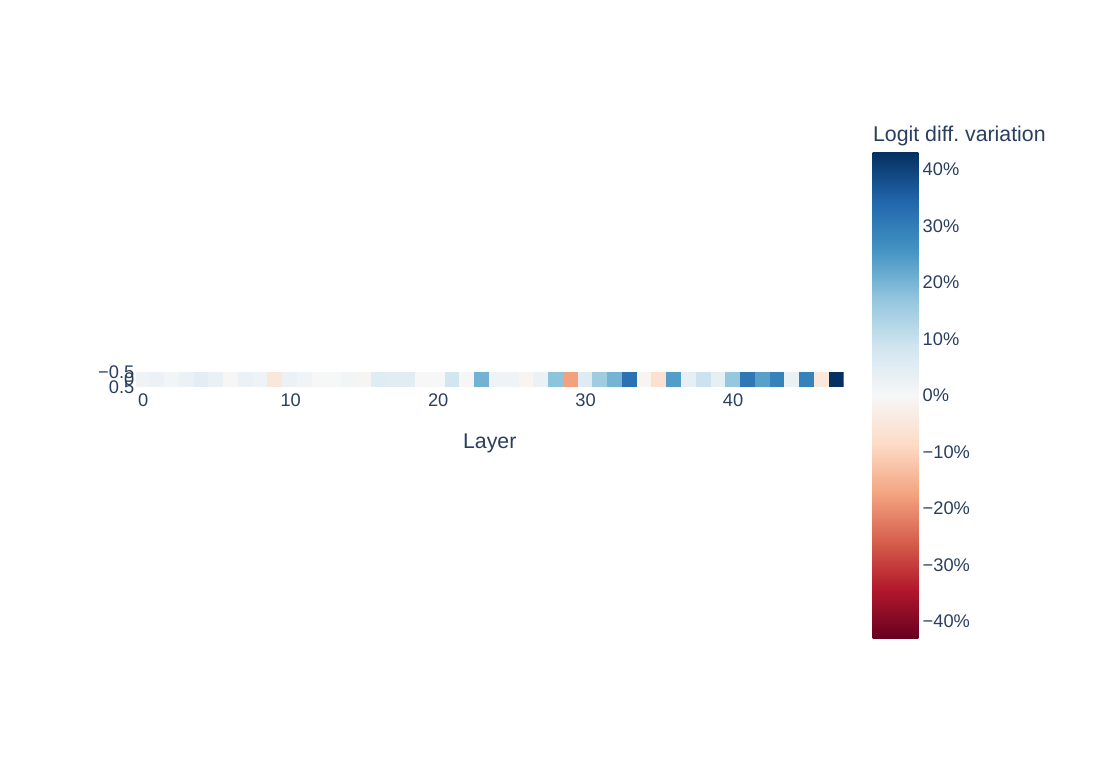}
    \caption{Direct effects of attention heads and MLPs for GPT-2 XL across syllogism tasks.}
    \label{gpt2xl}
\end{figure*}

\begin{figure*}[h]
    \centering
    \includegraphics[width=0.45\textwidth]{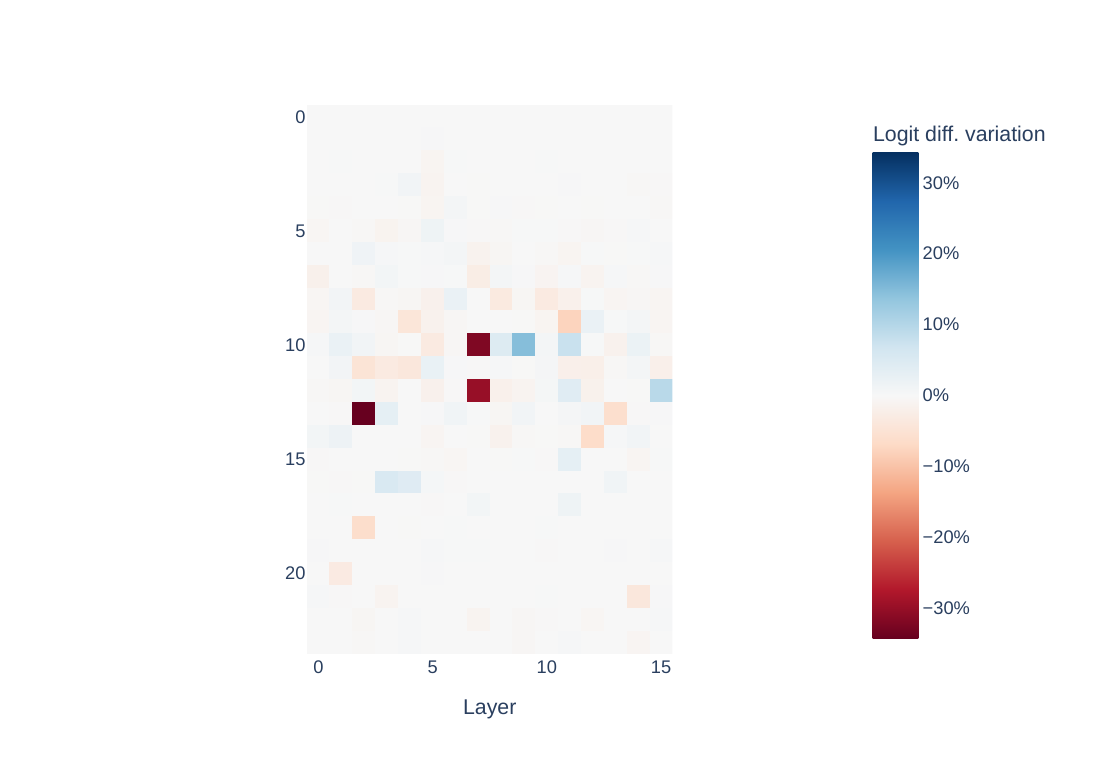}
    \includegraphics[width=0.45\textwidth]{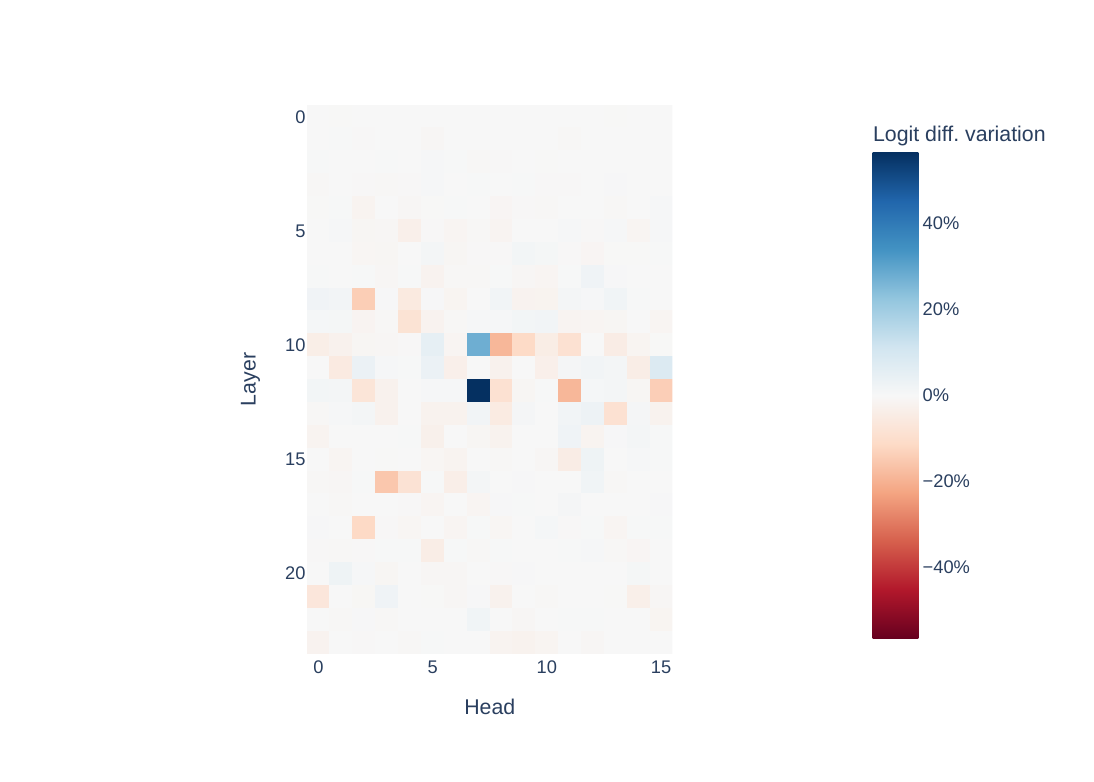}
    
    \includegraphics[width=0.45\textwidth]{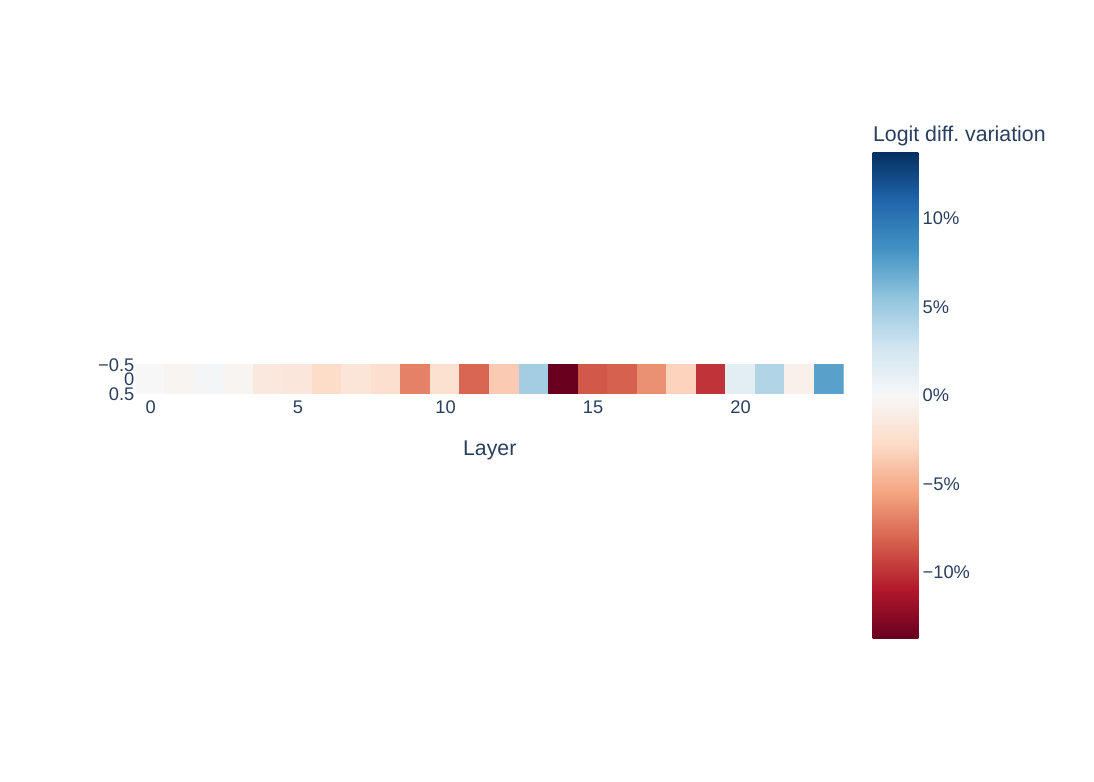}
    \includegraphics[width=0.45\textwidth]{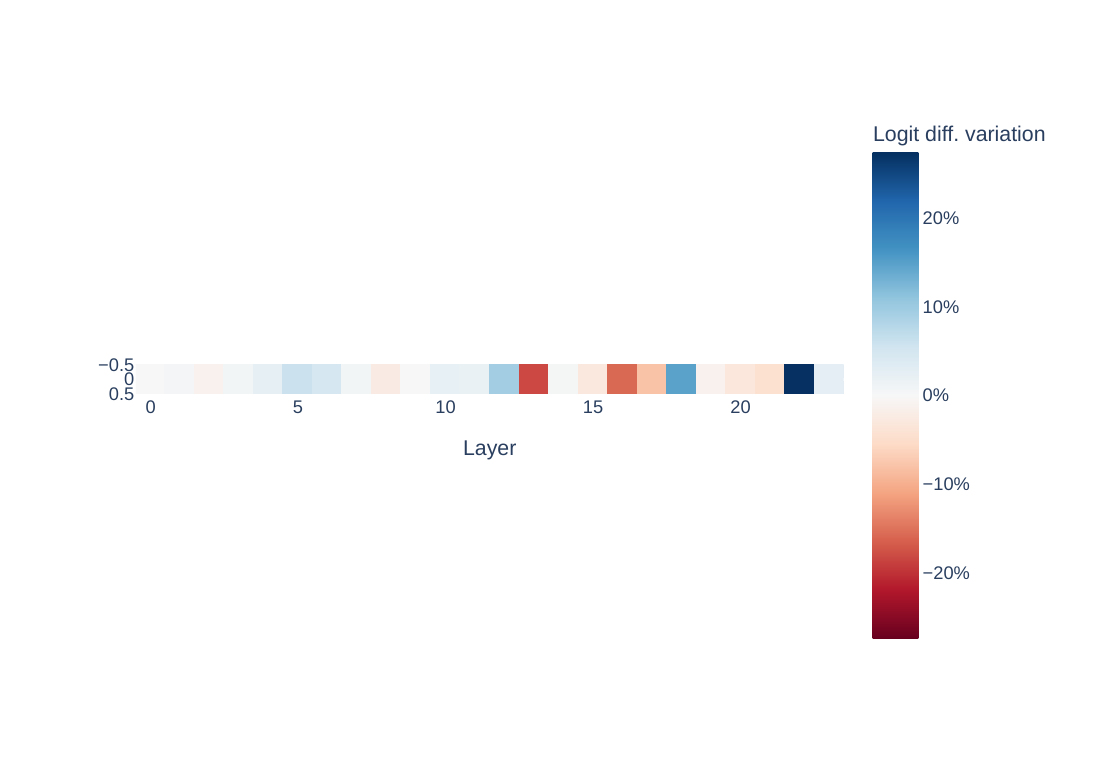}
    \caption{Direct effects of attention heads and MLPs for Pythia 1.4B across syllogism tasks.}
    \label{pythia}
\end{figure*}

\begin{figure*}[h]
    \centering
    \includegraphics[width=0.45\textwidth]{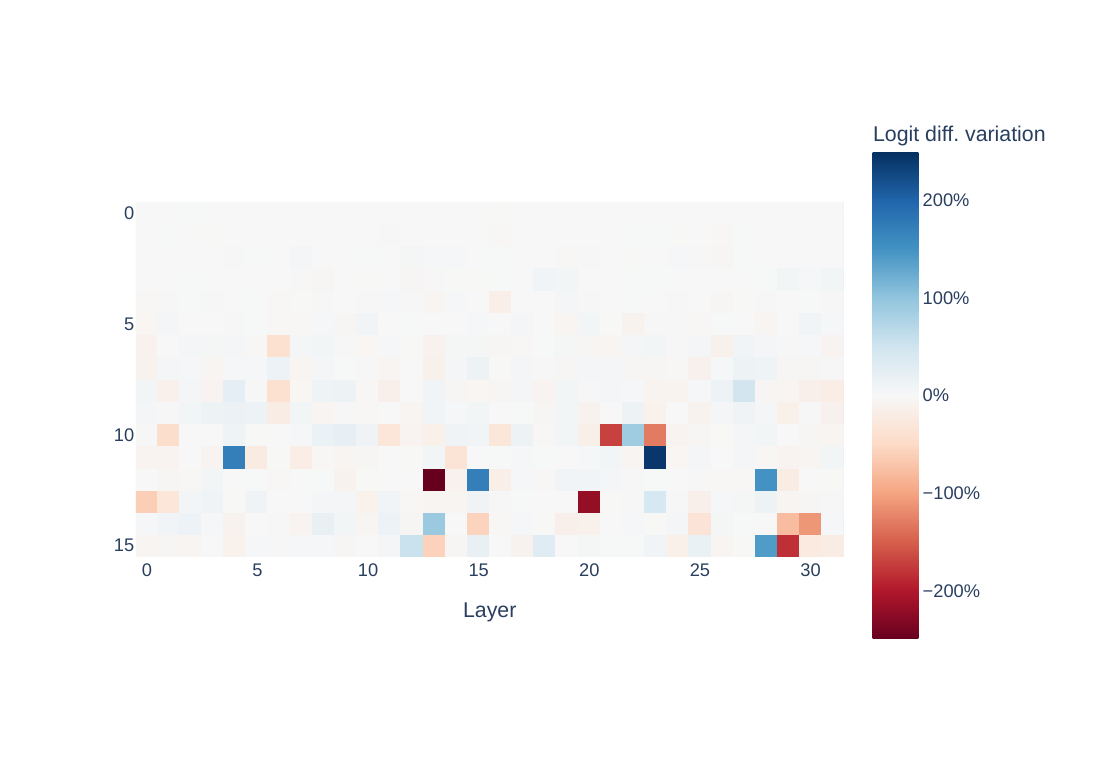}
    \includegraphics[width=0.45\textwidth]{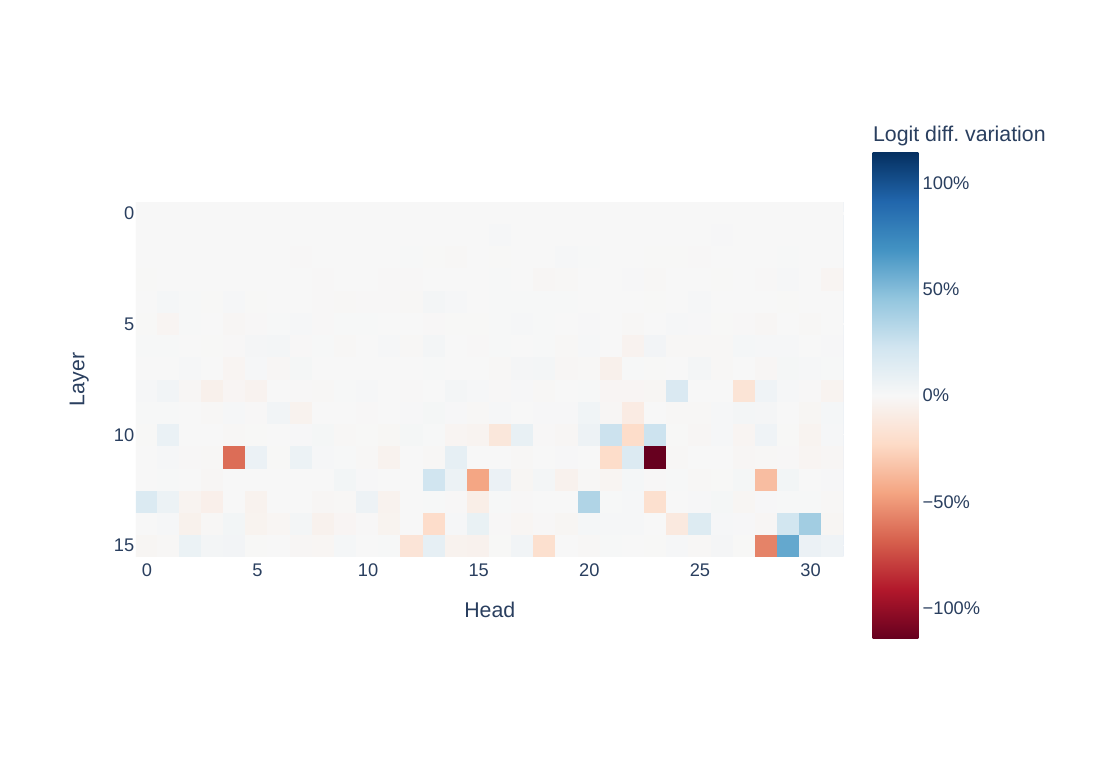}
    
    \includegraphics[width=0.45\textwidth]{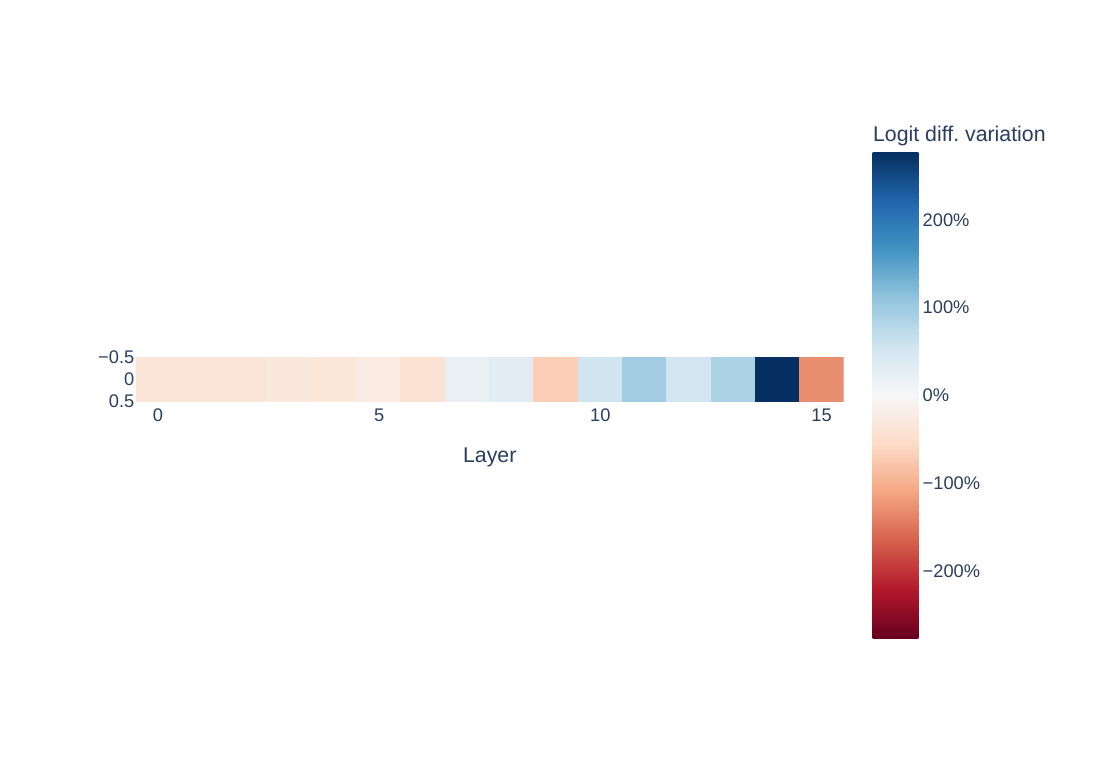}
    \includegraphics[width=0.45\textwidth]{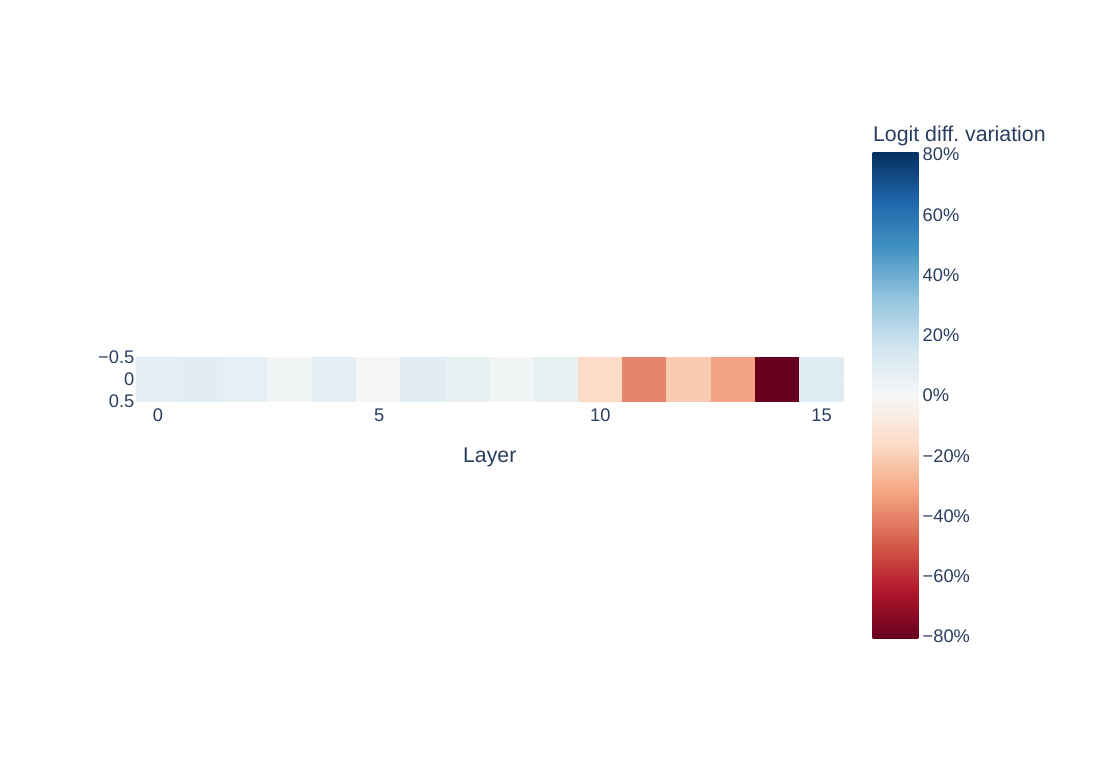}
    \caption{Direct effects of attention heads and MLPs for LLaMA 3.2B across syllogism tasks.}
    \label{llama3.2}
\end{figure*}

\begin{figure*}[h]
    \centering
    \includegraphics[width=0.45\textwidth]{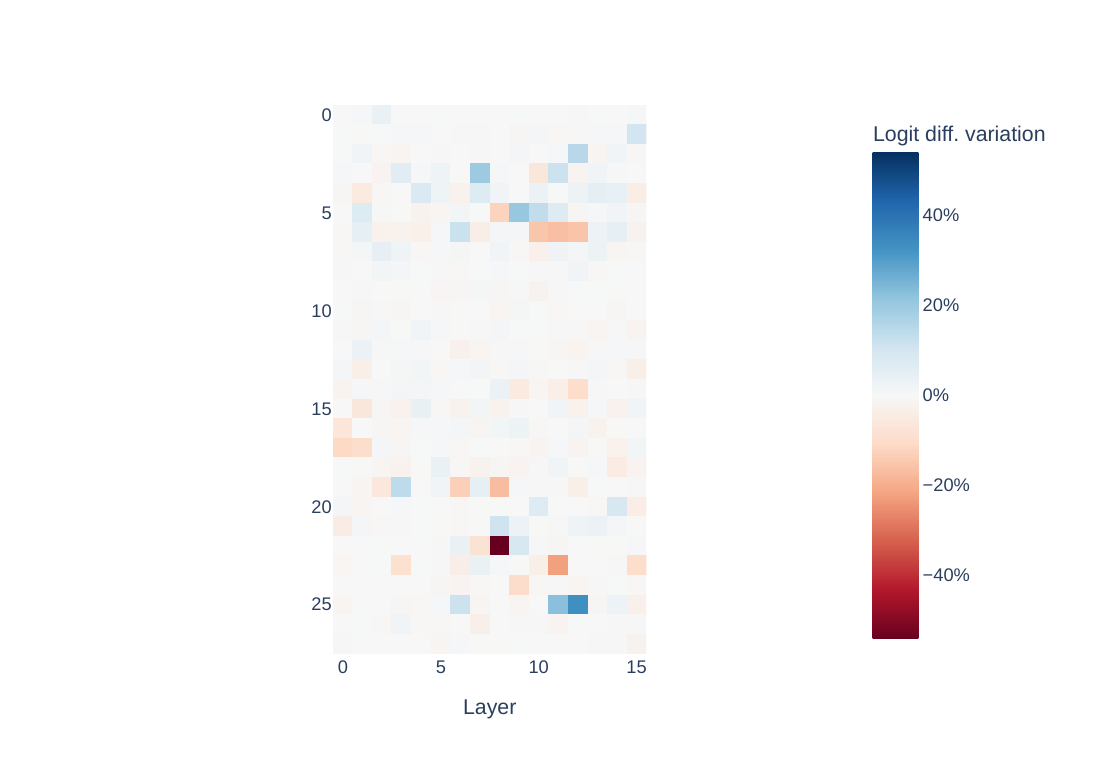}
    \includegraphics[width=0.45\textwidth]{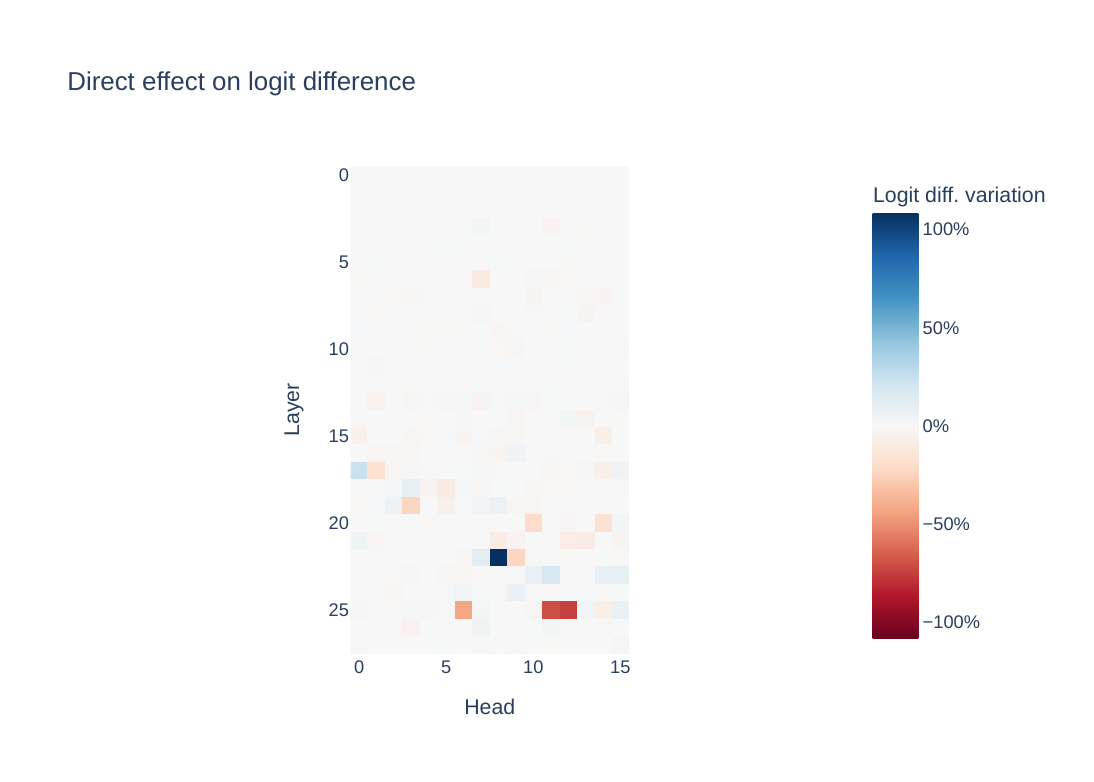}
    
    \includegraphics[width=0.45\textwidth]{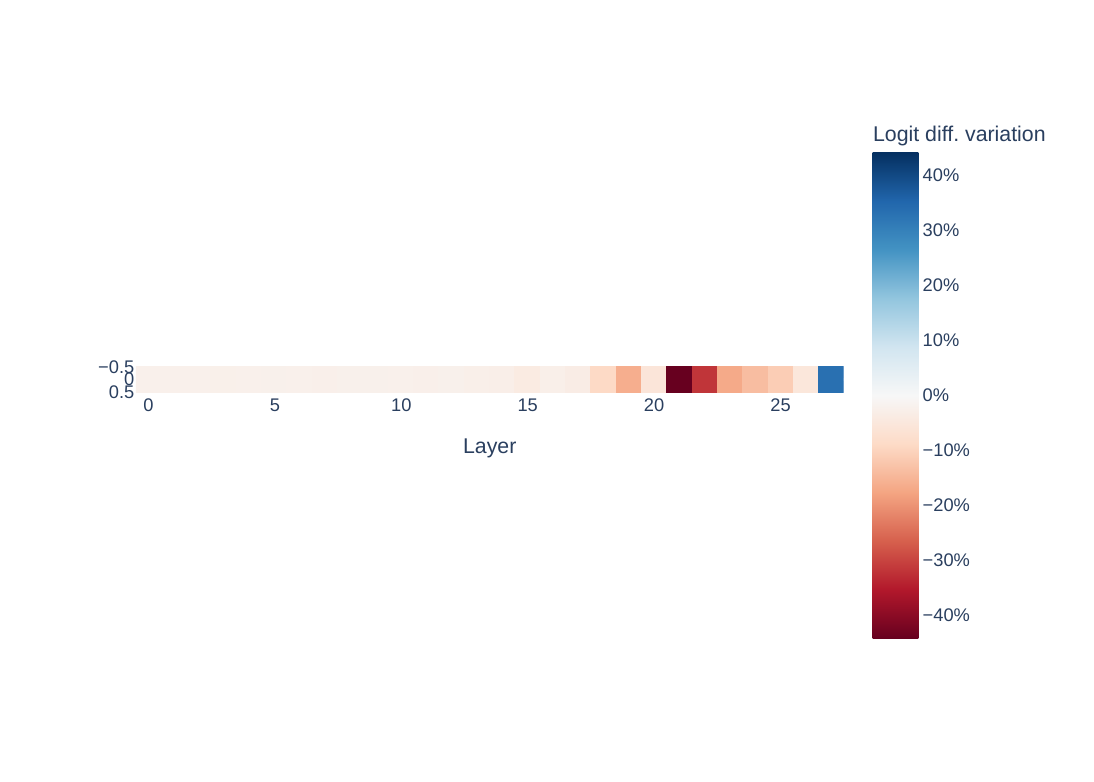}
    \includegraphics[width=0.45\textwidth]{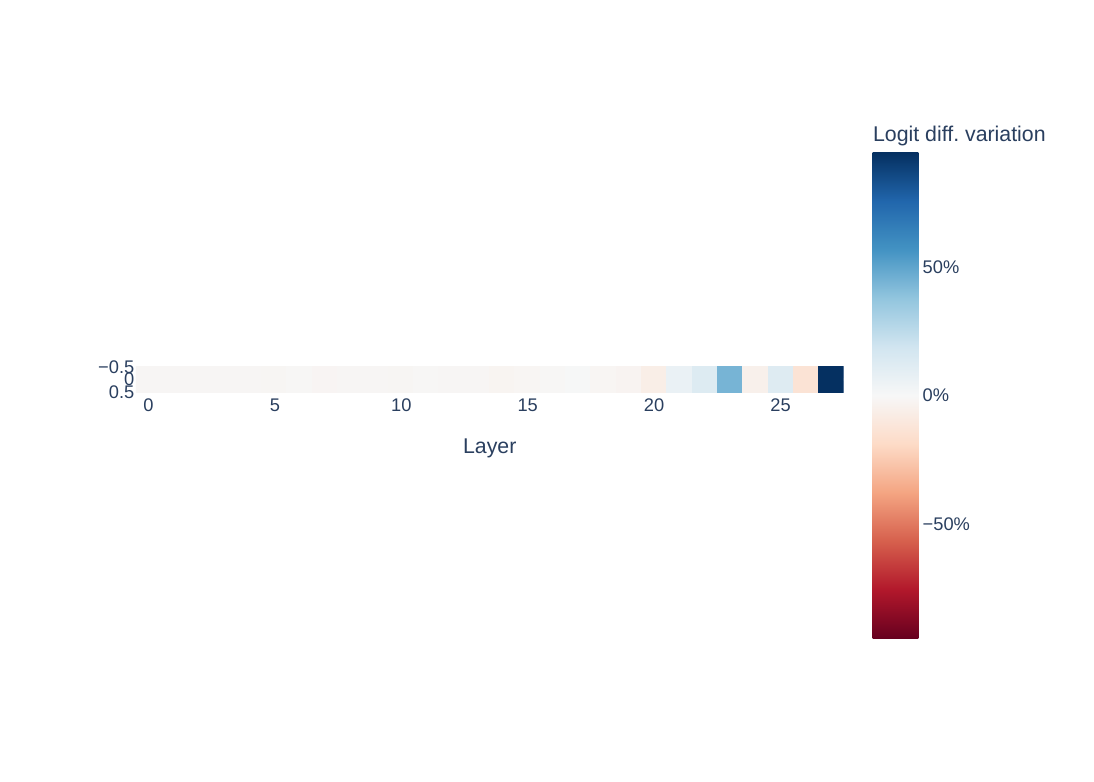}
    \caption{Direct effects of attention heads and MLPs for Qwen 1.7B across syllogism tasks.}
    \label{qwen}
\end{figure*}

These findings remain empirical and exploratory. Figures \ref{gpt2xl}-\ref{qwen} illustrate the direct effects of attention heads and MLPs across the syllogism tasks. A deeper investigation into how architectural scale affects circuit behavior remains a promising direction for follow-up work.

\begin{table*}[ht]
\centering
\begin{tabularx}{\textwidth}{Xcccc}
\toprule
& \textbf{GPT-2 XL} & \textbf{Qwen3-1.7B} & \textbf{LLaMA 3.2-1B} & \textbf{Pythia 1.4B} \\
\midrule
\textbf{Simple Syllogism} & 0.1112 & 0.5322 & $-0.4357$ & 1.0105 \\
\textbf{Opposite Syllogism} & 2.6114 & 1.5257 & $-0.1807$ & 2.1098 \\
\bottomrule
\end{tabularx}
\caption{Average logit difference across models and tasks.}
\label{ald_across_models}
\end{table*}

\end{document}